\documentclass[times, review, 10pt]{elsarticle}

\usepackage{graphicx}%
\usepackage{multirow,verbatim}%
\usepackage{amsmath,amssymb,amsfonts,amsthm}%
\usepackage{mathrsfs}%
\usepackage[title]{appendix}%
\usepackage{xcolor}%
\usepackage{textcomp}%
\usepackage{manyfoot}%
\usepackage{booktabs,enumitem}%
\usepackage{algorithm}%
\usepackage{algorithmicx}%
\usepackage{algpseudocode}%
\usepackage{listings}%
\usepackage{tikz}%
\usepackage{pgfplots}
\pgfplotsset{compat=newest}
\usepackage{pgfplotstable}
\usepackage{subcaption}
\usepackage{placeins}
\usepackage{siunitx}
\usepackage{rotating}
\usepackage{makecell}
\usepackage{url}
\usepackage[notransparent]{svg}
\usepgfplotslibrary{groupplots}
\usepackage{geometry}
\usetikzlibrary{backgrounds, calc}
\usetikzlibrary{positioning, decorations.pathreplacing, shapes.geometric, arrows.meta, fit}

\def\tsc#1{\csdef{#1}{\textsc{\lowercase{#1}}\xspace}}
\tsc{WGM}
\tsc{QE}

\newcommand{\norm}[1]{\left\lVert #1 \right\rVert}

\journal{Pattern Recognition}

\begin{document}

\begin{frontmatter}



\title{Epipolar-Constrained Graph Representation Learning for Geometric Pose Estimation}


\author[inst1]{Prateeth Rao}
\ead{prateeth.rao@iiitb.ac.in}

\author[inst1]{Sachit Rao\corref{cor1}}
\ead{sachit@iiitb.ac.in}

\cortext[cor1]{Corresponding author}

\affiliation[inst1]{organization={International Institute of Information Technology Bangalore},
    addressline={26/C, Electronics City},
    city={Bengaluru},
    postcode={560100},
    state={Karnataka},
    country={India}}

\begin{abstract}
We investigate geometry-aware graph representation learning for relative camera pose estimation using epipolar-constrained graph structures. Relative camera pose is commonly estimated from matched image correspondences that satisfy the epipolar constraint; however, identifying geometrically consistent correspondences under varying imaging conditions remains challenging. We employ an existing detector-free transformer architecture to establish correspondences between image pairs and construct a correspondence graph whose topology is guided by epipolar geometry. Initially, each node is connected to its (k) nearest neighbours based on spatial distance. The resulting graph is subsequently pruned using an initial estimate of the Essential Matrix, distinguishing the proposed formulation from conventional graph representations based primarily on heuristic spatial neighbourhoods. The graph learning module aggregates relational information between nodes into a graph-level representation that encodes the geometric context of the correspondences. This representation is passed to a pose regression network to estimate the relative camera rotation and translation. The model is trained across standard and wide baselines using a weighted objective that combines pose regression, scale consistency, heading consistency, and spectral properties of the pose parameters. Experimental graph analyses demonstrate that epipolar-guided edge pruning reduces graph connectivity by more than 90\% while retaining all correspondence nodes. Further spectral analyses characterize the effects of graph sparsification and injected outliers on learned graph representations. To investigate the influence of graph learning architectures, four variants of Graph Neural Networks (GNNs), namely Graph Convolutional Networks (GCN), Graph Attention Networks (GAT), Graph Isomorphism Networks (GIN), and Graph-based Cross Attention, are evaluated within a unified learning framework. The framework is evaluated on outdoor benchmark datasets, including KITTI, ETH3D, King's College, and TartanAir. Experimental results demonstrate consistent improvements in relative pose and trajectory estimation across multiple evaluation metrics compared with classical consensus-based estimators and learning-guided correspondence methods.
\end{abstract}


\end{frontmatter}

\section{Introduction}\label{sec:intro}

The existence of matched correspondences that satisfy certain geometric constraints across multiple views forms the fundamental basis of geometric reasoning in computer vision, enabling a wide range of applications including Visual SLAM, Structure-from-Motion (SfM), image registration, panoramic stitching, depth estimation, object pose estimation, and 3-D reconstruction. While recent advances in feature matching, either detector-based or detector-free, have significantly improved the concentration and robustness in finding image correspondences, these correspondences are also often contain redundant, noisy, and geometrically inconsistent. Detector-free methods, such as LoFTR, \cite{sun2021loftr}, directly learn from extracted visual features by modeling visual similarity across image pairs; while such approaches have substantially improved matching performance, they still produce large correspondences whose relational structure and geometric consistency remain difficult to exploit effectively for downstream vision tasks. Graphs form a modeling paradigm where such relational structures can be encoded naturally. In the context of monocular visual odometry (VO), a graph-based representation of matched correspondences has the potential to enhance relative pose estimation, thus determining camera motion, and at the same time, also increase the robustness of broader geometric vision systems that rely on accurate camera motion estimation; such downstream tasks include pose graph optimization, bundle adjustment, loop closure, dense mapping, and multi-view 3-D reconstruction.

In this paper, given numerous matched correspondences, each correspondence is selected as a node and those that are spatially or geometrically compatible, are connected using edges. Unlike treating matches independently, such a representation enables the consistency of a correspondence to be evaluated in the context of its neighboring matches, that is, geometrically consistent nodes exhibit compatible relative configurations between similar neighbours nodes, whereas incorrect matches introduce inconsistencies within their local neighborhood. GNNs provide a suitable framework for learning these relationships through message passing, allowing information from neighboring correspondences to be aggregated to refine their feature representations. This enables the network to reason jointly about correspondence appearance and the structural context in which each match occurs. However, the effectiveness of such relational reasoning is inherently dependent on the graph topology, since the choice of neighboring correspondences determines which relationships are exposed to the network. Consequently, constructing a graph that captures meaningful geometric relationships while avoiding spurious connections is essential for effectively exploiting the structure of dense correspondence sets.

To meet this requirement, we propose a geometry-aware epipolar correspondence graph that explicitly incorporates multi-view geometric constraints into graph construction. Instead of relying solely on heuristic neighborhood definitions, graph connectivity is established by preserving geometrically consistent correspondences while suppressing redundant and inconsistent relationships through epipolar constraints. The resulting sparse graph provides a compact yet informative representation that facilitates effective message passing and relational learning. We investigate the effectiveness of this representation across multiple GNN architectures and evaluate it on the task of relative camera pose estimation, as required in VO. 

\subsection{Related Work}

We briefly review other graph-based representations of a variety of matched correspondences, or features, in the literature. With the emergence of deep learning, feature detection and description became increasingly data-driven through methods such as SuperPoint, \cite{detone2018superpoint}; KeyNet, \cite{barroso2019key}; and, DISK \cite{disk-fua-2020}. Graph-based matchers such as SuperGlue, \cite{sarlin2020superglue}, and LightGlue, \cite{lindenberger2023lightglue}, introduced contextual reasoning between detected features. More recently, detector-free approaches such as COTR, \cite{jiang2021cotr}; LoFTR \cite{sun2021loftr}; ASpanFormer, \cite{chen2022aspanformer}; and ROMA, \cite{edstedt2024roma}, directly establish correspondences between image regions without requiring an explicit keypoint detection stage. These developments have shifted feature matching from sparse, hand-engineered correspondence estimation toward dense and learned matching frameworks that are more suitable for challenging conditions involving viewpoint changes, illumination variations, sparse textures, and wide-baseline observations. As these approaches are predominantly trained on large-scale datasets containing diverse image pairs and geometric variations, the learned correspondence relations capture spatial and contextual dependencies that are difficult to model using handcrafted descriptors. Recent learning-based approaches have explored attention and transformer architectures to capture local and global geometric context for robust correspondence estimation and outlier suppression \cite{LinLLHANet2026,LiuTransMatch2025,WangCHCANet2025,SongTransBoNet2024}. Semantic-guided coarse-to-fine matching and epipolar constraints have further been investigated to improve geometric consistency in feature matching and optical flow estimation \cite{ZhangAreaPoint2026,FanEpipolar2024}. Similarly, geometric and monocular depth priors have been incorporated into multi-view stereo and visual odometry frameworks, alongside dense optical flow and sequence-level optimization for improved pose estimation \cite{TangMCMVSNet2026,FuDepthPrior2026,LiuFSOVO2026}.

Graph-based methods explicitly propagate information between feature nodes through message passing or attention mechanisms, enabling contextual relationships between potential correspondences to be learned. Detector-free methods similarly exploit self- and cross-attention over dense feature representations to establish correspondences across images, while geometric supervision, including depth, camera pose, optical flow, or epipolar relationships, further guides correspondence learning. Consequently, the resulting matches are not solely based on local appearance similarity but implicitly or explicitly incorporate the geometric consistency between the two views. This makes learned correspondence methods particularly attractive for visual odometry, where reliable geometrically consistent matches are essential for subsequent relative pose estimation.

Graph-based visual perception methods have explored different representations of spatial, feature, and correspondence relationships across images and 3-D observations. These include keypoint and image patch based graph representation \cite{jin2019drone,elmoogy2021pose}, spatial area and point cloud based neighbourhoods \cite{srivastava2021exploiting}, and correspondence graphs constructed using $k$-NN, Delaunay triangulation, positional encodings, and bipartite relationships \cite{imgkeygnnxu2022,pautrat2023gluestick,lindenberger2023lightglue}. Recent approaches have further explored hierarchical clustering, joint 2-D--3-D neighbourhoods, adaptive edge connectivity, and multimodal correspondence graphs \cite{gyamfi2024hierarchical,dgcgnnwang2024,2026GIMS}, reflecting a shift towards more flexible graph representations.

Learning over constructed graphs has employed a range of GNN, attention, and matching-based architectures depending on the underlying task. Early GNN approaches used keypoint relation graphs for pose estimation, \cite{jin2019drone}, while Graph Convolution Network (GCN)-based models directly regressed global pose vectors from graph representations, \cite{elmoogy2021pose}. Relational graphs have also been learned using combinations of GNNs and MLPs \cite{srivastava2021exploiting}. Attention-based methods have been used specifically for correspondence learning. These approaches have incorporated increasingly complex attention and message-passing mechanisms, including self- and cross-attention with heterogeneous message passing and point-to-point and line-to-line cost computation, \cite{pautrat2023gluestick}; and attention between nodes of a main graph and a seeded graph \cite{seededchen2021}. Hierarchical methods propagate information from local nodes to centroids and between centroids for coarse-to-fine matching \cite{gyamfi2024hierarchical}, while adaptive approaches employ edge-conditioned message passing, multi-head attention, and optimal transport, \cite{2026GIMS}. Other methods combine graph convolution with geometric and colour features followed by 2-D--3-D cross-graph attention, \cite{dgcgnnwang2024}; GCN/GAT layers with differentiable Sinkhorn matching, \cite{imgkeygnnxu2022}, or self- and cross-attention with optimal transport and dustbin handling for unmatched points, \cite{sarlin2020superglue}. More recent formulations further incorporate adaptive GNN depth and progressive pruning using Sinkhorn-based matching scores, \cite{lindenberger2023lightglue}. Transformer-based methods further combine graph propagation with global attention, MLP-based feature mixing, and structured fusion to improve representation capacity while reducing computational cost, \cite{zheng2025sct,zhong2026sdformer}.

\subsection{Contributions}

Most existing graph construction strategies primarily encode local proximity, feature similarity, or correspondence availability, without explicitly enforcing underlying geometric constraints. Such constructions can therefore produce graphs that are locally meaningful but geometrically ambiguous, leaving the subsequent learning process to discover the relevant camera geometry from the data. For relative camera pose estimation, this is particularly restrictive because corresponding observations across two images are not arbitrary: they are constrained by the epipolar geometry futher induced by chirality constraints of positive depth. Incorporating this constraint during graph construction can therefore provide a stronger inductive bias by explicitly connecting observations that are geometrically consistent across views, rather than relying solely on spatial or feature-based proximity. An epipolar graph, as we propose, consequently allows the learning model to operate on a representation that already reflects the underlying multi-view geometry, potentially improving robustness to noisy correspondences and reducing the burden on the network to implicitly learn geometric relationships from unconstrained graph connectivity. 

\begin{figure*}[htpb!]
    \centering
    \includegraphics[width=0.8\linewidth]{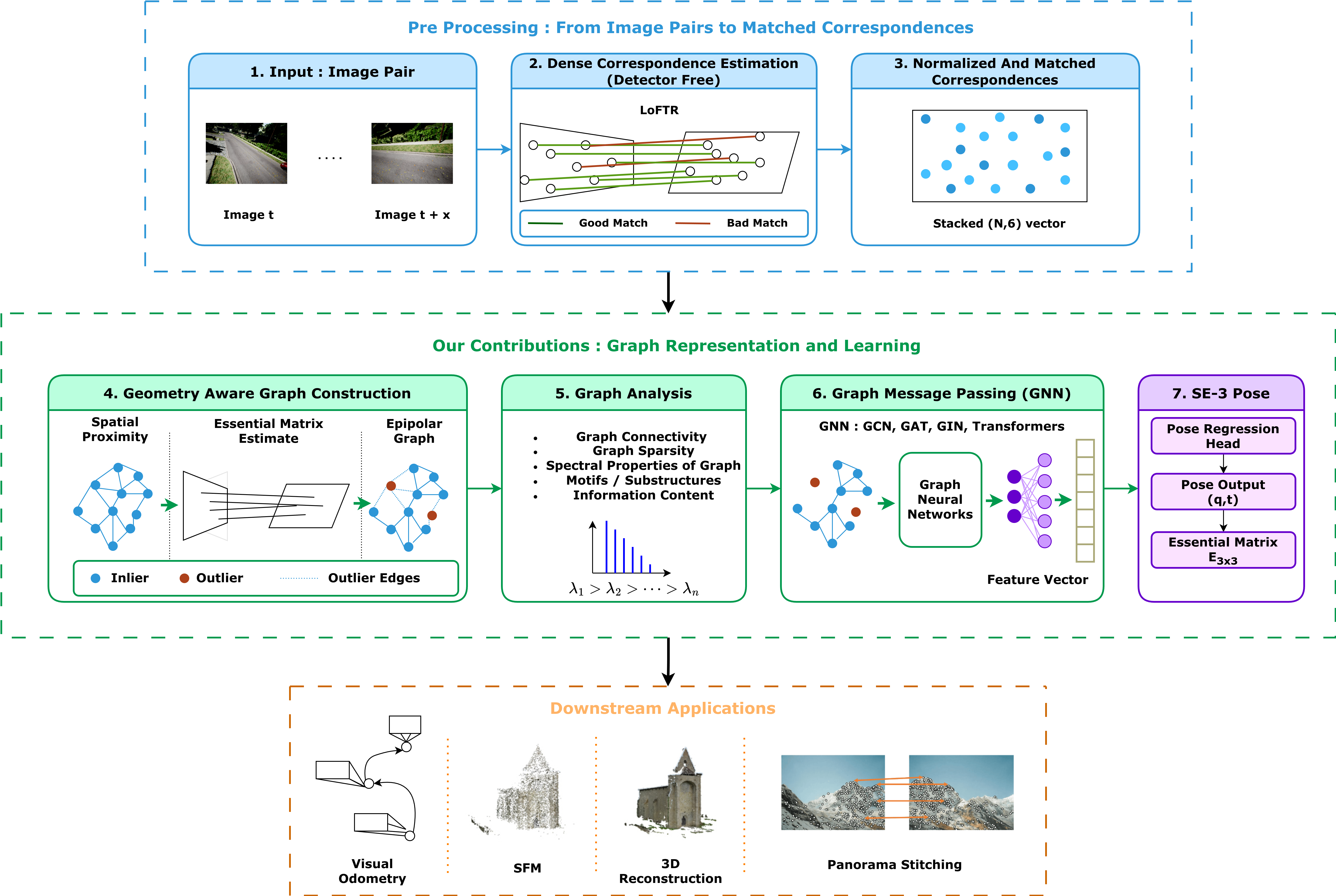}
    \caption{A General Block Diagram of our pipeline that includes: a) Pre Processing - Stereo Initialization to Matched Correspondences, b) Our Contributions - Graph Construction, Graph Analysis, Graph Learning and Camera Pose Estimation, and c) Downstream Applications.}
    \label{fig1:graph-block-diag}
\end{figure*}

In the context of learning, current approaches rely on multiple convolution, message-passing, or attention layers to progressively discover relationships within the graph. For geometric pose estimation, however, a carefully constructed graph already provides a strong relational structure, particularly when its connectivity incorporates geometric constraints such as epipolar relationships. The sparse nature of such graphs means that meaningful information can propagate through a relatively small number of local connections without requiring deep stacks of attention or message-passing operations. Consequently, a lightweight combination of GNN layers and a compact regression module can potentially capture the relevant local and global structure while reducing computational and memory overhead. This is particularly advantageous for low-resolution and sparse geometric graphs, where the objective is not to learn arbitrary relationships between all observations, but to exploit the structured relationships that have already been embedded during graph construction. The block diagram of our proposed approach is shown in Fig.~\ref{fig1:graph-block-diag}.

Compared with conventional VO and SLAM pipelines (such as SVO, DSO, MonoSLAM, LSD-SLAM) or learning-based relative pose estimation (such as RPNet, VLocNet++, DirectionNet, DiffPoseNet), joint learning of correspondences and camera pose provides an opportunity to reduce the dependence on separately optimized intermediate modules. Instead of independently detecting features, establishing correspondences, estimating geometric constraints, and subsequently refining the pose, a learned model can exploit the relationships between observations and pose within a unified framework. This can reduce error propagation between successive stages and allow the model to learn robust representations for challenging conditions such as sparse textures, illumination changes, and wide-baseline motion. In particular, combining correspondence information with geometric pose estimation retains the advantages of explicit image-to-image relationships while avoiding the need to rely entirely on handcrafted geometric optimization or a sequence of independently tuned VO/SLAM components; see \cite{forster2014svo, engel2017direct, davison2007monoslam, engel2014lsd} for conventional pipelines and \cite{en2018rpnet, chen2021wide, parameshwara2022diffposenet} for learning-based approaches.

For the VO application, the contributions of the paper are:
\begin{enumerate}[label=\arabic*.,wide, labelindent=0pt]

    \item \textbf{Geometry-aware correspondence graph construction and analysis}: We propose an epipolar correspondence graph that integrates spatial proximity with projective geometric constraints, enabling compact and geometrically consistent graph topologies. We demonstrate that geometry-aware edge selection yields more informative relational representations than conventional heuristic graph construction strategies.

    \item \textbf{Graph representation learning for correspondence reasoning}: We investigate how geometry-aware graph representations facilitate correspondence reasoning across multiple GNN architectures. By comparing graph learning models under a common representation, we demonstrate that improved graph topology enables more effective aggregation of geometrically consistent correspondences while suppressing noisy matches.

    \item \textbf{Geometry-coupled supervision for graph learning}: We develop a composite training objective that jointly constrains relative camera pose, Essential Matrix consistency, and geometric scale, and demonstrate that coupling these complementary objectives yields more robust graph representation learning than individual supervision terms.

    \item \textbf{Evaluation through relative camera pose estimation}: We validate the proposed graph representation on the correspondence-driven task of relative camera pose estimation within a monocular visual odometry pipeline. Extensive experiments demonstrate consistent improvements across geometric pose evaluation metrics, highlighting the effectiveness of geometry-aware graph construction for downstream geometric reasoning.

\end{enumerate}

The paper is organised as follows: In Sec.~\ref{sec:prelims}, the epipolar constraint for pixels in normalised coordinates is introduced and the motivation to employ graph-based learning is highlighted; basic graph-learning concepts are also introduced. The proposed methodology is presented in Sec.

\section{Preliminaries}\label{sec:prelims}
This section outlines the foundational theoretical concepts required for our proposed methodology. We first present the geometric principles governing two-view geometry and camera calibration via epipolar geometry. Subsequently, we formalize graph representation learning techniques used to model structural dependencies in matched correspondences that are determined using a detector-free matching algorithm.

\subsection{Feature-matching}\label{sec:FeatMatch}

We employ the LoFTR algorithm \cite{sun2021loftr}, to obtain matched correspondences in a pair of images. While any matching algorithm can be employed, for the datasets considered in this paper, this algorithm yielded better-suited matches than other algorithms. This algorithm computes fine-grained matches and a confidence score for each match. We select a fixed set of $n$ matched keypoint correspondences per image pair based on top confidence scores. The normalised pixel coordinates for all $n$ matched pairs of correspondences in all $(N-1)$ pairs of images form the geometric input representation for subsequent relative pose and graph embedding modules.

\subsection{Epipolar Geometry}\label{subsec:epipolar_geometry}

Let $\mathbf{X}_i = \begin{bmatrix}x_i & y_i & z_i & 1\end{bmatrix}^T, \ i=1,2$, be two matched correspondences, from images $I_1$ and $I_2$. respectively, expressed in 3-D world coordinates. Using the camera projection matrix $\mathbf{P}$, \cite{HartleyZisserman2004}, they can be projected to a 2-D image plane in pixel coordinates $\mathbf{x}_i=\begin{bmatrix}u_i & v_i & 1\end{bmatrix}^T$, using the relation $\mathbf{x}_i = \mathbf{P} \mathbf{X}_i$, where $\mathbf{P} = \mathbf{K}\begin{bmatrix}\mathbf{R} & \mathbf{t}\end{bmatrix}$, $\mathbf{K} \in \mathbb{R}^{3 \times 3}$ is the camera intrinsic matrix, and $\mathbf{R} \in SO(3)$ and $\mathbf{t} \in \mathbb{R}^3$ denote the rotation and translation of the camera between images $I_1$ and $I_2$. The matrix $\mathbf{K}$ is of the form $\mathbf{K} = \begin{bmatrix} f_x & s & c_x \\ 0 & f_y & c_y \\ 0 & 0 & 1 \end{bmatrix}$, where $f_x, f_y$ represent focal lengths, $(c_x, c_y)$ is the principal point, and $s$ is the axis skew. The pixel coordinates $\mathbf{x}_i$ are normalized to 2-D homogeneous coordinates, using $\hat{\mathbf{x}}_i=\mathbf{K}_1^{-1}\mathbf{x}_i$.

The goal of VO is to estimate $\mathbf{R}, \mathbf{t}$ between the two frames, but using only the matched correspondences, $\hat{\mathbf{x}}_{1,2}$, from the corresponding images that are available. $\mathbf{R}, \mathbf{t}$ can be found by calculating the Essential Matrix (EM), $\mathbf{E} \in \mathbb{R}^{3 \times 3}$, from the matched correspondences, \textit{if} the pair also satisfy the epipolar constraint
\begin{equation}\label{EpiConst}
    \hat{\mathbf{x}}_2^T \mathbf{E} \hat{\mathbf{x}}_1 = 0, \ \hat{\mathbf{x}}_i=\begin{bmatrix}\hat{u}_i & \hat{v}_i & 1\end{bmatrix}^T.
\end{equation}
The EM plays a crucial role since it can be decomposed into $\mathbf{E} = [\mathbf{t}]_{\times} \mathbf{R}$, where $[\mathbf{t}]_{\times}$ is the skew-symmetric matrix representation of the vector $\mathbf{t}$. Further, from the Singular Value Decomposition of $\mathbf{E}=\mathbf{U} \mathbf{\Sigma} \mathbf{V}^T$, and defining the orthogonal matrix $\mathbf{W} = \begin{bmatrix} 0 & -1 & 0 \\ 1 & 0 & 0 \\ 0 & 0 & 1 \end{bmatrix}$, the ``correct'' $\mathbf{R}, \mathbf{t}$ are one of four possible structural solutions $\mathbf{R} \in \{\mathbf{U} \mathbf{W} \mathbf{V}^T, \, \mathbf{U} \mathbf{W}^T \mathbf{V}^T\}, \quad \mathbf{t} \in \{+\mathbf{u}_3, \, -\mathbf{u}_3\}$, where $\mathbf{u}_3$ is the third column of $\mathbf{U}$. The physically valid pose is selected by performing a cheirality test to ensure that the 3-D points lie strictly in front of both cameras (positive depth). Note that the EM is a rank-2 matrix, with identical non-zero eigenvalues; thus $\mathbf{\Sigma} = \text{diag}(1, 1, 0)$.

Now, given the ``right'' (which satisfy \eqref{EpiConst}) $\left(\hat{\mathbf{x}}_1,\hat{\mathbf{x}}_2\right)$ pairs, algorithms, such as the Eight-point algorithm, \cite{Hartley1997}, can be implemented to find $\mathbf{E}$ (as the name suggests, 8 unique pairs are required). Finding this unique set amongst several matched pairs is the main challenge - we propose using graph-based representation and learning to address it. The feasibility of this approach is now introduced. 

Given images $I_1,I_2$, let there be $j\gg1$ pairs of matched correspondences, $\left(\hat{\mathbf{x}}_{1j},\hat{\mathbf{x}}_{2j}\right),\ j =1,2,\cdots$, where the set $k=\{1,2,\cdots\}, \ |k|\leq j$ of pairs also satisfy the epipolar constraint; as a result, \eqref{EpiConst} can also be rewritten as the weighted equality
\begin{equation}\label{EpiConstSum}
    \sum\limits_{w\in k}\left(\alpha_w\hat{\mathbf{x}}_{2w}^T\right)\mathbf{E} \left(\alpha_w\hat{\mathbf{x}}_{1w}\right) = 0,
\end{equation}
where the weights $0<\alpha_w\leq 1$. Further, let each of the vectors $\left(\hat{\mathbf{x}}_{1w},\hat{\mathbf{x}}_{2w}\right)$ be mapped to a different vector space, even a higher-dimensional one, but using the same mapping operator; such an operation also preserves the epipolar constraint. 

With this background, we propose and expect graph-learning, after message passing between connected epipolar nodes and pooling, to filter the ``right'' set of pairs in some higher-dimensional space. First, the matched correspondences generated by the keypoint matching algorithm that are close to each other when measured by Euclidean distance of pixels are connected using edges. Next, this graph is pruned based on an initial estimate of the EM. This estimate is used to calculate the Sampson distance, \cite{Sampson1982}, between matched correspondences, $\left(\hat{\mathbf{x}}_{1w},\hat{\mathbf{x}}_{2w}\right)$, given by 
\begin{equation} \label{sampson_dist}
    d_{\text{Sampson}}\left(\hat{\mathbf{x}}_{1w}, \hat{\mathbf{x}}_{2w}\right) = \frac{\left(\hat{\mathbf{x}}_{2w}^T \mathbf{E} \hat{\mathbf{x}}_{1w}\right)^2}{\norm{\mathbf{E}\hat{\mathbf{x}}_{1w}}^2 + \norm{\mathbf{E}^T\hat{\mathbf{x}}_{2w}}^2},
\end{equation}
which in turn, is used as a basis for pruning. As can be observed from \eqref{sampson_dist}, in the ideal case, if $\left(\hat{\mathbf{x}}_{1w},\hat{\mathbf{x}}_{2w}\right)$ are truly epipolar, then, from \eqref{sampson_dist}, the Sampson distance $d_{\text{Sampson}}\left(\hat{\mathbf{x}}_{1w}, \hat{\mathbf{x}}_{2w}\right)=0$. Graph-learning is introduced next.

\subsection{Graph Representation Learning}\label{subsec:graph_representation_learning}

A Keypoint Correspondence Graph, denoted by $\mathcal{G} = (\mathcal{V}, \mathcal{E}, \mathbf{W})$, is a weighted and undirected graph constructed from matched correspondences, where $\mathcal{V} = \{v_1, v_2, \cdots, v_n\}$ is the set of $n$ keypoint correspondences between consecutive frames; $\mathcal{E} \subseteq \mathcal{V} \times \mathcal{V}$ is the set of edges based on $k$-NNs; and $\mathbf{W}: \mathcal{E} \rightarrow \mathbb{R}^+$ contains the assignment of edge weights based on geometric consistency; the choice of $k$ is clarified later. The node feature vector is constructed using the matched correspondences in a sequence of $N$ input images $\mathcal{I} = \{I_1, I_2, \dots, I_N\}$. Given $(N-1)$ consecutive image pairs $(I_t, I_{t+1})$ for $t \in \{1, \dots, (N-1)\}$ and the pair $p=1,\cdots,(N-1)$, the feature of node $v_i^p$ in the graph, denoted by $\mathbf{h}_i^{p} \in \Re^6$, is given by
\begin{equation}
    \mathbf{h}_i^{p} = \begin{bmatrix} \hat{\mathbf{x}}_{t,i}^{p} & \hat{\mathbf{x}}_{(t+1),i}^{p}\end{bmatrix}^T, \ t,p=1,\cdots,(N-1), \ i=1,\cdots,n.
\end{equation}
An initial spatial proximity graph $\mathcal{G}_0^{p} = (\mathcal{V}_0^{p}, \mathcal{E}_0^{p})$, is constructed by connecting each node $v_i^p$ to its $k$-nearest neighbors ($k$-NN) based on Euclidean distances in the 2-D image plane. 

The implementation of any feature matching algorithm on any pair of images may result in $\mathcal{G}_0^{p}$ to have many vertices and edges, where, not all elements of the feature vector of a node satisfy the epipolar constraint. Thus, this graph is pruned by removing geometrically invalid edges based on evaluating the Sampson distance for each node $v_i^p\in V_0^{p}$. The pruned graph is denoted as the \textit{Epipolar Graph}, $\mathcal{G}_{\text{Epi}}^{p} = \left(\mathcal{V}^{p}, \mathcal{E}_{\text{Epi}}^{p}\right)$, where $\mathcal{E}_{\text{Epi}}^{p}$ consists of edges $e_{ij} \in \mathcal{E}_0^{p}$, which connect node pairs $v_i,v_j$, such that the Sampson distance satisfies $d_{\text{Sampson}}\left(\hat{\mathbf{x}}_{t,w}^{p}, \hat{\mathbf{x}}_{(t+1),w}^{p}\right)< \tau, \ w=i,j$, that is, it is within a specified threshold - a small number. All other edges in the initial graph are eliminated. Thus, the resulting graph, which may be sparse, is expected to capture true underlying geometric constraints. Note that edge pruning may break $\mathcal{G}_{\text{Epi}}^{p}$ into multiple disconnected components or isolated nodes.

Rather than discarding isolated nodes resulting from edge pruning, the entire set of nodes $\mathcal{V}^{p}$ (with or without incident edges) is retained along with the modified adjacency structure $\mathcal{E}_{\text{Epi}}^{p}$. This complete graph structure is passed directly to the Graph Representation Learning module to preserve full feature coverage. The Epipolar Graph $\mathcal{G}_{\text{epi}}^{p}$ with initial node features $\mathbf{H}^{(0)} \in \Re^{n \times 6}$ ($n$ is the number of matched pairs, Sec.~\ref{sec:FeatMatch}) is processed through GNNs to aggregate structural and geometric dependencies. Regardless of the underlying architecture, all model variants follow a unified functional pipeline:
\begin{equation}\label{GraphPipeline}
    \mathcal{G}_{\text{Epi}}^{p} \xrightarrow{\quad\text{Graph Model}\quad} \mathcal{RG}^{p} \xrightarrow{\quad\text{Pooling}\quad} \mathbf{z}^{p}
\end{equation}
$\mathcal{G}_{\text{Epi}}^{p}$ is mapped by a Graph Model into an enriched Representation Graph, $\mathcal{RG}^{p}$, containing contextualized node features $\mathbf{H}^{L} \in \Re{R}^{n' \times d_z}$, where, subsequently, a global graph-level pooling operation condenses these node representations into a single fixed-dimensional embedding vector, $\mathbf{z}^{p} \in \Re^{d_z}$. The architectures proposed in this paper are described in Sec.~\ref{sec:GxnArch}.

Broadly, when the feature vectors of all input nodes are such that they satisfy the epipolar constraint, the graph architecture is expected to maintain this relation even at the output. In this ideal case, as the variation in the nodes will be small, the graph is thus expected to act as a ``low-pass'' filter and not distort the epipolar constraint. However, as this ideal situation cannot and does not hold, as the correspondence matching algorithm will provide matches that do not satisfy the epipolar constraint, such matches become outliers or anomalies. As is known, in such cases, the graph is expected to act as a ``high-pass'' filter. Instead of tuning graph parameters to act as either of these filters, we explore the joint use of convolutional and attention mechanisms.

Attention mechanisms connect every node to every other node, but such that their combination is weighted, thus, if two nodes, where one is an ``inlier'' and the other is an anomaly, the weighted contribution of the anomalous node is small. A single convolutional layer, on the other hand, propagates information between immediate neighbours, and thus an inlier and outlier neighbour pair can get ``smoothened'' out. Thus, one of the architectures we propose (which also performs better than others) is a combined GCN-GAT architecture, using which we aim to strike a balance between handling anomalies as well as truly matched correspondences. We also design the number of convolutional layers based on the properties of the pruned graph $\mathcal{G}_{\text{Epi}}^{p}$, such as the vertex degree. We also present results using the Graph Isomorphism Network (GIN), where the in-built MLP is expected to handle such variations.

\subsection{Relative Camera Pose Regression with Geometric Learning}\label{subsec:pose_regression}

The pooled global feature vector $\mathbf{z}^{p}$ summarizes the spatial and geometric structural context of the correspondence graph. This representation is passed to an MLP to regress relative camera motion parameters: a unit orientation quaternion $\mathbf{q}_{\text{Pr}} \in \Re^4$ and a translation vector $\mathbf{t}_{\text{Pr}} \in \Re^3$.

With this background, the main results of the paper are presented next. 

\section{Main Results}

The training loss, error metrics to compare performance with other algorithms, and implementation details of the proposed Variants are presented in this section. To reiterate, the input to the pipeline, as shown in Fig.~\ref{fig1:graph-block-diag}, is an image pair, $I_t,I_{(t+1)}$, from which the Epipolar Graph is constructed. The output are the predicted pose parameters, denoted by $\left(\mathbf{q}_{\text{Pr}}, \mathbf{t}_{\text{Pr}}\right)$. To apply this pipeline to the VO problem, $\mathbf{q}_{\text{Pr}}$ is converted the rotation matrix $\mathbf{R}_{\text{Pr}} \in SO(3)$ and together with $\mathbf{t}_{\text{Pr}}$, the predicted EM is constructed as $\mathbf{E}_{\text{Pr}} = \mathbf{t}_{\text{Pr}\times}\mathbf{R}_{\text{Pr}}$. $\mathbf{R}_{\text{Pr}}$ and $\mathbf{t}_{\text{Pr}}$ are further cast into the $4 \times 4$ rigid-body transformation matrix between frames $t,(t+1)$
\begin{equation}
    \mathbf{T}_{t, (t+1)} = \begin{bmatrix} \mathbf{R}_{\text{Pr}} & \mathbf{t}_{\text{Pr}} \\ \mathbf{0}^T & 1 \end{bmatrix}
\end{equation}
The absolute camera pose at frame $(t+1)$ is recovered by recursively chaining relative frame transformations, to obtain $\mathbf{T}_{\text{Abs}}^{(t+1)} = \mathbf{T}_{\text{Abs}}^{t} \mathbf{T}_{t, (t+1)}$; the matrix $\mathbf{T}_{\text{Abs}}^{t}$ is initialised by setting $\mathbf{R}_{\text{Pr}}$ as the identity matrix and $\mathbf{t}_{\text{Pr}}$ as the vector of zeros. By performing this procedure for all $p=1,\cdots,(N-1)$ frames, the VO problem is solved. 

\subsection{Loss Function for Training}

To enforce orientation consistency, directional translation alignment, and algebraic properties of the EM, the multi-component geometric training loss $\mathcal{L}_{T}$ is formulated as:
\begin{equation}\label{LossEq}
    \mathcal{L}_{T} = \lambda_{P}\left(\mathcal{L}_{\mathbf{q}} + \mathcal{L}_{td} + \mathcal{L}_{s}\right) + \lambda_{F}\mathcal{L}_{E}^{F} + \lambda_{SVD}\mathcal{L}_{E}^{SVD} + \lambda_{\psi}\mathcal{L}_{\psi}
\end{equation}
where $\lambda_{P}, \lambda_{F}, \lambda_{SVD},$ and $\lambda_{\psi}$ are scalar balancing hyper-parameters. The individual loss terms are defined as follows:

\begin{enumerate}[label=\textit{\roman*}.,wide, labelindent=0pt]
    \item \textbf{Quaternion Loss}: $\mathcal{L}_{\mathbf{q}}=\|\tilde{\mathbf{q}}-\mathbf{q}_{GT}\|$, where $\mathbf{q}_{GT}$ is the ground-truth (GT) quaternion. Denoting the raw network prediction as $\mathbf{q}_{Pr}$, the aligned term $\tilde{\mathbf{q}}$ resolves the antipodal ambiguity:
    \begin{equation}\label{qpred}
        \tilde{\mathbf{q}}= \begin{cases}
            \mathbf{q}_{Pr} & \text{if } \mathbf{q}_{Pr}^T\mathbf{q}_{GT} \geq 0 \\
            -\mathbf{q}_{Pr} & \text{otherwise}.
        \end{cases}
    \end{equation}
    Since both $\pm\mathbf{q}$ represent identical rotations, forcing sign alignment avoids gradient step discontinuities caused by non-unique hemispherical coverage during optimization.

    \item \textbf{Translation Direction and Scale Loss}: These components enforce geometric alignment in direction and scale and are given by
    \begin{subequations}\label{TSLoss}
        \begin{align}
            \mathcal{L}_{td} &= 1 - \frac{\mathbf{t}_{Pr}}{\|\mathbf{t}_{Pr}\|} \cdot \frac{\mathbf{t}_{GT}}{\|\mathbf{t}_{GT}\|} \label{TdLoss} \\
            \mathcal{L}_{s} &= \left|\|\mathbf{t}_{Pr}\|-\|\mathbf{t}_{GT}\|\right|, \label{TsLoss}
        \end{align}
    \end{subequations}
    where $\mathbf{t}_{Pr}$ and $\mathbf{t}_{GT}$ denote predicted and GT translation vectors, respectively.

    \item \textbf{EM Structural Loss}: Combines a Frobenius reconstruction term $\mathcal{L}_{E}^{F}=\|\mathbf{E}_{Pr}-\mathbf{E}_{GT}\|_F$ with an algebraic constraint derived from singular value properties of valid EMs, given by $\mathcal{L}_{E}^{SVD} = (\sigma_1 - \sigma_2)^2 + \sigma_3^2$, where $\sigma_{1,2,3}$ are the singular values of $\mathbf{E}_{Pr}$. It is highlighted that Valid EMs strictly satisfy $\sigma_1 = \sigma_2>0$ and $\sigma_3 = 0$.

    \item \textbf{Heading Angle (Yaw) Loss:} $\mathcal{L}_{\psi}$ constrains the yaw component extracted from relative rotation to enforce directional trajectory consistency along planar navigation axes.
\end{enumerate}

\subsection{Proposed Architectures}\label{sec:GxnArch}

As shown in \eqref{GraphPipeline}, we explore four architectural strategies to create $\mathcal{RG}^{p}$:
\begin{enumerate}
    \item \textbf{Variant A:} GCN layers $\rightarrow$ GAT layer $\rightarrow$ Global Average Pooling
    \item \textbf{Variant B:} GAT layer $\rightarrow$ GCN layers $\rightarrow$ Global Average Pooling
    \item \textbf{Variant C:} GIN Module $\rightarrow$ Global Sum Pooling
    \item \textbf{Variant D:} Graph Cross-Attention Module $\rightarrow$ Global Average Pooling
\end{enumerate}

For Variants A, B, and C, the graph $\mathcal{G}_{\text{Epi}}^{p}$ is processed directly through the respective graph network to yield $\mathcal{RG}^{p} = \text{GraphModel}\left(\mathcal{G}_{\text{Epi}}^{p}, \mathbf{H}^{(0)}\right) \in \Re^{n \times d_{z}}$. For Variant D, the epipolar keypoints are decomposed into two distinct graphs: $\mathcal{G}_1^{p}$ with node features from image $I_t$ and $\mathcal{G}_2^{p}$ with node features from image $I_{(t+1)}$. The Graph Cross-Attention module models directional feature exchange between these two view representations to yield $\mathcal{RG}^{p} = \text{CrossAttn}\left(\mathcal{G}_1^{p}, \mathcal{G}_2^{p}\right) \in \Re^{n \times d_{z}}$. The resulting node representations $\begin{bmatrix} \mathbf{h}_1 & \cdots & \mathbf{h}_{n} \end{bmatrix}^T$ are aggregated to form $\mathbf{z}^{p} \in \Re^{d_{z}}$. Variants A, B, and D employ Global Average Pooling (GAP), $\mathbf{z}^{p} = \dfrac{1}{n} \sum\limits_{i=1}^{n} \mathbf{h}_i$, while Variant C utilizes Global Sum Pooling, $\mathbf{z}^{p} = \sum\limits_{i=1}^{n} \mathbf{h}_i$, to retain multiset structural scale.

Hyperparameters governing graph construction and network depth were determined empirically through validation on the KiTTi dataset. To construct the Epipolar Graph, $\mathcal{G}_{\text{Epi}}^{p}$, the local neighborhood graph topology uses $k = 6$ for $k$-NN connectivity. Edges with epipolar misalignment exceeding the Sampson error threshold $\tau = 10^{-4}$ are pruned. A summary of the graph properties before and after these steps of pruning are presented in Table~\ref{tab:sec5-sub2-epi-graph1}; the SEQ01 and SEQ09 sequences in the KiTTi dataset are used to generate these tables.

\begin{table*}[htbp]
  \centering
  \caption{Graph Metrics Comparison }
  \label{tab:sec5-sub2-epi-graph1}
  \small
  \begin{tabular}{l *{6}{r}}
    \toprule
		\multicolumn{7}{c}{\textbf{Temporal Gap = 1 frame}}\\
		\midrule
    & \multicolumn{2}{c}{\textbf{Before ($k$-NN)}} & \multicolumn{2}{c}{\textbf{After (Epipolar)}} & \multicolumn{2}{c}{\textbf{Change ($\Delta$)}} \\
    \cmidrule(lr){2-3} \cmidrule(lr){4-5} \cmidrule(lr){6-7}
    \textbf{Metric} & \textbf{SEQ01} & \textbf{SEQ09} & \textbf{SEQ01} & \textbf{SEQ09} & \textbf{SEQ01} & \textbf{SEQ09} \\
    \midrule
    Edges & 11,448 & 18,516 & 2,356 & 933 & -9,092 (94.82\%) & -17,583 (96.45\%) \\
    Average Degree & 6.00 & 6.00 & 1.24 & 0.30 & -4.76 & -5.70 \\
    Density (Average) & 0.0031 & 0.0019 & 0.0006 & 0.0001 & -0.0025 & -0.0018 \\
    GT Inliers Preserved (\%) & 100.00 & 100.00 & 7.54 & 5.08 & -92.46 & -94.92 \\
		\midrule
		\multicolumn{7}{c}{\textbf{Temporal Gap = 5 frames}} \\ 
		\midrule
		Edges & 3,138 & 6,180 & 513 & 407 & -2,625 (91.43\%) & -5,773 (94.83\%) \\
    Average Degree & 6.00 & 6.00 & 0.98 & 0.40 & -5.02 & -5.60 \\
    Density (Average) & 0.0115 & 0.0058 & 0.0019 & 0.0004 & -0.0096 & -0.0054 \\
    GT Inliers Preserved (\%) & 100.00 & 100.00 & 10.08 & 5.39 & -89.92 & -94.61 \\
		\midrule
		\multicolumn{7}{c}{\textbf{Temporal Gap = 10 frames}} \\
		\midrule
		Edges & 1,938 & 2,274 & 304 & 168 & -1,634 (81.78\%) & -2,106 (89.09\%) \\
    Average Degree & 6.00 & 6.00 & 0.94 & 0.44 & -5.06 & -5.56 \\
    Density (Average) & 0.0186 & 0.0159 & 0.0029 & 0.0012 & -0.0157 & -0.0147 \\
    GT Inliers Preserved (\%) & 100.00 & 100.00 & 22.46 & 8.38 & -77.54 & -91.62 \\
    \bottomrule
  \end{tabular}
\end{table*}

With the initial estimate of the EM and the resulting Sampson error threshold, for consecutive image pairs (temporal gap = 1) from the SEQ01 and SEQ09 sequences in the KiTTi dataset, close to 95\% of the edges in the $k$-NN graph removed by geometric filtering, thus creating a sparse graph where only geometrically consistent connections remain; the inliers are reduced to a mere 7\% of the original set. This result is also an indicator of the number of outliers in the matched correspondences that would have to be handled by a graph architecture had this pruning not been performed. The reduction in diameter and shortest path indicates a more compact, semantically meaningful graph structure. The average degree after filtering stabilizes at $0.3-1.2$, indicating sparse but connected components.

As the temporal gap increases (upto 10 frames gap), the node count decreases from between 1900--3000 to 320--380 due to fewer correspondences. However, edge retention increases from about 5\% to 18--22\%, suggesting more stable geometric relationships.  

Based on these observations, for Variant A, we employ $L = 3$ stacked GCN layers followed by 1 multi-head GAT layer. The rationale is that for the pruned graph with an effective average degree $\bar{k}' \in [0.4, 1.2]$, three GCN layers provide sufficient receptive field, where Layer 1 aggregates direct geometric neighbors (validated matches), Layer 2 propagates information across connected components, and Layer 3 integrates global context through hub nodes (high-degree vertices). For Variant B (GAT-GCN cascade), we utilize 1 multi-head GAT layer followed by $L = 2$ GCN layers. We experiment with changing the ordering of the attention and convolution modules in learning representations. 

The results of implementing these architectures on several datasets are presented next. 

\section{Implementation Results}\label{sec:experimental_setup}

To evaluate the proposed Graph Representation Learning framework rigorously for relative camera pose regression, we design a comprehensive experimental protocol. This section outlines the selected outdoor benchmark datasets, data partitioning strategies, hardware-software training environments, and parameter choices governing graph construction and network training.

\subsection{Datasets and Data Partitioning}\label{subsec:datasets}

Our evaluation focuses exclusively on outdoor environments, which present distinct geometric challenges including wide baseline variations, dynamic lighting, scale drift, and non-uniform keypoint distributions. We evaluate our pipeline across four diverse outdoor benchmarks:
\begin{itemize}
    \item \textbf{KiTTi 2012 SLAM Benchmark:} Serves as the primary dataset for initial graph construction, model prototyping, and baseline validation under real-world autonomous driving conditions.
    \item \textbf{ETH3D Outdoor Benchmark:} Offers high-precision ground-truth camera poses across varied outdoor scenes with challenging perspective shifts.
    \item \textbf{Tartan Air (TA) (Neighborhood Sequence):} Provides photorealistic synthetic trajectories with diverse lighting, weather conditions, and challenging camera dynamics.
    \item \textbf{King's College (KC) Dataset:} A large-scale urban outdoor scene from the Cambridge Landmarks benchmark designed for wide-baseline visual localization.
\end{itemize}

Since intrinsic camera parameters fluctuate across different sequences, thus altering keypoint projection geometries onto the image plane, data splitting is performed strictly on an individual sequence basis rather than mixing frames across sequences. For every sequence across all benchmarks: the \textbf{first 200 consecutive frames} are reserved exclusively for evaluation and testing to preserve natural sequence trajectories and the remaining frames of the sequence are partitioned using a 75:25 ratio for training and validation, respectively.

\subsection{Implementation Details}\label{subsec:implementation_details}

The end-to-end framework is implemented in \texttt{PyTorch 2.6} leveraging \texttt{PyTorch Geometric (PyG)} for graph-structured operations. Model optimization and benchmarking are executed on Kaggle cloud compute infrastructure using distinct GPU environments. Model training is conducted using an NVIDIA Tesla P100 GPU (16\,GB VRAM), whereas model evaluation and inference benchmarking are executed on an NVIDIA Tesla T4 GPU. Models are trained using the Adam optimizer with an initial learning rate of $\eta = 10^{-4}$. We apply a \texttt{ReduceLROnPlateau} scheduler to dynamically attenuate the learning rate upon validation loss stagnation. Training is performed with a batch size of $8$ image pairs, while evaluation batch size is set to $16$. Models are trained for a maximum of $100$ epochs, incorporating an early stopping criterion with a patience of $25$ epochs based on validation performance.

\subsection{Performance on Evaluation Metrics}\label{sec:evalmet}

To provide a comprehensive assessment, evaluation is categorized into pose estimation accuracy and downstream 3-D reconstruction quality; the following metrics are reported. 

\begin{enumerate}[label=\textit{\roman*}.,wide, labelindent=0pt]
    \item \textbf{Discernible Rotation Error (DRE) and Discernible Translation Error (DTE)}, \cite{lee2024s}: By denoting $\mathbf{R}_{Pr}$ and $\mathbf{R}_{GT}$ as the predicted and ground-truth rotation matrices, respectively, these errors are given by 
    \begin{equation}\label{DTEDREEq}
			\mathrm{DRE} = \cos^{-1}\left(\frac{\mathrm{Tr}(\mathbf{R}_{GT}^T\mathbf{R}_{Pr})-1}{2}\right) \ \text{and} \ \mathrm{DTE} = \cos^{-1}\left(\frac{\mathbf{t}_{Pr}^T\mathbf{t}_{GT}}{\|\mathbf{t}_{Pr}\|\|\mathbf{t}_{GT}\|}\right). 
		\end{equation}
    Note that the translation error is expressed as an angular measure. To perform DRE, the quaternions $\mathbf{q}_{Pr}$ estimated from the architecture is expressed in the form of the rotation matrix, $\mathbf{R}_{Pr}$.

    \item \textbf{Absolute Pose Error (APE)}: consists of translational and rotational components, given by
    \begin{equation}\label{APEGenEEq}
			\mathrm{APE}_t^{(k)} = \|\mathbf{p}_{Pr,k}-\mathbf{p}_{GT,k}\|_2 \ \text{and} \ \mathrm{APE}_r^{(k)} = \cos^{-1}\left(\frac{\mathrm{Tr}(\mathbf{R}_{err,k}) - 1}{2}\right),
		\end{equation}
    where $\mathrm{R}_{err,k} = (\mathbf{R}_{GT,k})^T \mathbf{R}_{Pr,k}$ and $\mathbf{p}_{Pr,k},\mathbf{p}_{GT,k}$ are the predicted and GT positions at frame $k$, respectively.

    \item Global consistency is summarized by the Root Mean Square (RMS) of the \textbf{Absolute Trajectory Error (ATE)} over $N$ frames, given by
    \begin{equation}\label{ATEEq}
        \mathrm{ATE} = \sqrt{\frac{1}{N}\sum_{k=1}^{N} \|\mathbf{p}_{Pr,k}-\mathbf{p}_{GT,k}\|_2^2 }.
    \end{equation}

\end{enumerate}

As trajectory estimates often contain a mixture of small drift, scale ambiguity, and sporadic large deviations due to tracking loss or degenerate configurations. Since ATE relies on least-squares alignment, it is highly sensitive to such deviations; a few large failures can dominate the metric and obscure improvements in the majority of correctly estimated poses. DTE and DRE mitigate this issue through robust alignment and bounded residual aggregation. As a result, they remain sensitive to changes in the inlier error distribution, which is particularly important in VO where performance improvements typically manifest as reduced drift rather than elimination of all failure cases. Detailed experimental results using these metrics are presented next.

\begin{table*}[htbp]
  \centering
  \caption{Comparison of DTE and DRE metrics across models for the first 10 samples of the ETH3D Botanical Garden test dataset. The best (lowest) value for each metric per sample is highlighted in bold.}
  \label{tab:dte_dre_comparison}
  \small
  \resizebox{\textwidth}{!}{%
  \begin{tabular}{c | cc | cc | cc | cc | cc | cc | cc | cc | cc}
    \toprule
    & \multicolumn{8}{c|}{\textbf{Ours}} & \multicolumn{10}{c}{\textbf{Baselines}} \\
    \cmidrule(lr){2-9} \cmidrule(lr){10-19}
    & \multicolumn{2}{c|}{\textbf{GAT+2xGCN}} & \multicolumn{2}{c|}{\textbf{3xGCN+GAT}} & \multicolumn{2}{c|}{\textbf{GIN\_SumPool}} & \multicolumn{2}{c|}{\textbf{CrossGraph}} & \multicolumn{2}{c|}{\textbf{PoseNet}} & \multicolumn{2}{c|}{\textbf{RPNet}} & \multicolumn{2}{c|}{\textbf{RPNet+}} & \multicolumn{2}{c|}{\textbf{DiffPoseNet (C)}} & \multicolumn{2}{c}{\textbf{DiffPoseNet (F)}} \\
    \textbf{Sample} & \textbf{DTE} & \textbf{DRE} & \textbf{DTE} & \textbf{DRE} & \textbf{DTE} & \textbf{DRE} & \textbf{DTE} & \textbf{DRE} & \textbf{DTE} & \textbf{DRE} & \textbf{DTE} & \textbf{DRE} & \textbf{DTE} & \textbf{DRE} & \textbf{DTE} & \textbf{DRE} & \textbf{DTE} & \textbf{DRE} \\
    \midrule
    1  & \textbf{0.1152} & \textbf{6.71}  & 0.1166 & 15.07 & 0.6351 & 13.33 & 0.4299 & 23.06 & 0.4405 & 69.54 & 0.1758 & 40.44 & 0.1791 & 19.33 & 0.2410 & 79.54 & 0.2115 & 34.49 \\
    2  & 0.0966          & 22.11          & \textbf{0.0657} & \textbf{3.19}  & 0.5854 & 29.55 & 0.3179 & 15.91 & 0.5617 & 16.76 & 0.1792 & 115.72& 0.2501 & 9.62  & 0.5550 & 90.68 & 0.9087 & 21.14 \\
    3  & 0.1754          & 26.53          & \textbf{0.1465} & \textbf{6.13}  & 0.5869 & 36.18 & 0.2257 & 17.32 & 0.4963 & 25.87 & 0.2596 & 130.74& 0.3958 & 6.24  & 0.3022 & 53.05 & 0.9313 & 16.56 \\
    4  & \textbf{0.1592} & 26.31          & 0.1799 & 10.05          & 0.4848 & 27.87 & 0.4855 & 19.09 & 0.3875 & 5.38  & 0.2174 & 66.68 & 0.4977 & \textbf{5.26} & 0.6372 & 49.50 & 0.4826 & 7.11  \\
    5  & 0.2269          & 18.05          & \textbf{0.1966} & \textbf{2.42}  & 0.6671 & 22.23 & 0.2700 & 16.33 & 0.3896 & 31.54 & 0.2012 & 113.36& 0.3231 & 14.83 & 0.3864 & 45.54 & 0.6157 & 22.33 \\
    6  & \textbf{0.0924} & 19.47          & 0.1172 & \textbf{6.29}  & 0.5936 & 30.91 & 0.4562 & 16.94 & 0.4305 & 42.17 & 0.2552 & 64.53 & 0.4308 & 22.17 & 0.4035 & 47.30 & 0.8911 & 17.90 \\
    7  & 0.1922          & 28.78          & 0.1724 & 17.11          & 0.6098 & 53.33 & 0.2272 & 26.43 & 0.6878 & 20.19 & \textbf{0.0567} & 176.03& 0.5127 & \textbf{11.10} & 0.7581 & 82.46 & 0.6349 & 17.72 \\
    8  & 0.1691          & 30.25          & \textbf{0.1304} & \textbf{11.13} & 0.6460 & 30.34 & 0.2980 & 16.71 & 0.5387 & 31.22 & 0.2325 & 138.84& 0.6175 & 18.55 & 0.3407 & 70.71 & 0.9286 & 15.32 \\
    9  & 0.2057          & 25.25          & 0.1745 & \textbf{5.02}  & 0.5987 & 44.59 & 0.2148 & 18.95 & 0.4427 & 35.43 & \textbf{0.1651} & 81.09 & 0.6108 & 12.21 & 0.3118 & 43.39 & 0.2333 & 22.19 \\
    10 & 0.1352          & 31.07          & 0.1026 & 12.86          & 0.5786 & 37.57 & 0.2683 & 20.25 & 0.5006 & 26.62 & \textbf{0.0601} & 169.92& 0.5389 & \textbf{6.12}  & 0.1412 & 24.82 & 0.5024 & 7.72  \\
    \bottomrule
  \end{tabular}%
  }
\end{table*}

\begin{table*}[htbp]
  \centering
  \caption{Comparison of DTE and DRE metrics across architectures for the first 10 samples of KiTTi Sequence 01. The best (lowest) value for each metric per sample is highlighted in bold.}
  \label{tab:dte_dre_kitti}
  \small
  \resizebox{\textwidth}{!}{%
  \begin{tabular}{c | cc | cc | cc | cc | cc | cc | cc | cc | cc}
    \toprule
    & \multicolumn{8}{c|}{\textbf{Ours}} & \multicolumn{10}{c}{\textbf{Baselines}} \\
    \cmidrule(lr){2-9} \cmidrule(lr){10-19}
    & \multicolumn{2}{c|}{\textbf{GAT+2xGCN}} & \multicolumn{2}{c|}{\textbf{3xGCN+GAT}} & \multicolumn{2}{c|}{\textbf{GIN\_SumPool}} & \multicolumn{2}{c|}{\textbf{CrossGraph}} & \multicolumn{2}{c|}{\textbf{PoseNet}} & \multicolumn{2}{c|}{\textbf{RPNet}} & \multicolumn{2}{c|}{\textbf{RPNet+}} & \multicolumn{2}{c|}{\textbf{DiffPoseNet (C)}} & \multicolumn{2}{c}{\textbf{DiffPoseNet (F)}} \\
    \textbf{Sample} & \textbf{DTE} & \textbf{DRE} & \textbf{DTE} & \textbf{DRE} & \textbf{DTE} & \textbf{DRE} & \textbf{DTE} & \textbf{DRE} & \textbf{DTE} & \textbf{DRE} & \textbf{DTE} & \textbf{DRE} & \textbf{DTE} & \textbf{DRE} & \textbf{DTE} & \textbf{DRE} & \textbf{DTE} & \textbf{DRE} \\
    \midrule
    1  & 1.0025 & 2.5708 & 1.0021 & 2.8769 & 1.0018 & 2.3858 & 1.0042 & 2.5684 & 1.0118 & 4.3889 & 1.0022 & 136.4704 & 1.0079 & \textbf{2.3739} & \textbf{0.9995} & 15.4007 & 1.0180 & 2.4699 \\
    2  & 0.0039 & 0.6293 & 0.0039 & \textbf{0.5895} & 0.0247 & 2.1226 & \textbf{0.0027} & 0.9363 & \textbf{0.0024} & 2.8734 & 0.0039 & 161.9227 & 0.0085 & 1.1246          & 0.0056          & 18.2815 & 0.0206 & 3.4445 \\
    3  & 0.0030 & \textbf{0.3539} & \textbf{0.0027} & 0.4681 & 0.0221 & 2.2639 & 0.0036 & 0.6876 & 0.0134 & 2.1653 & 0.0028 & 134.0428 & 0.0113 & 0.6281          & 0.0034          & 13.5461 & 0.0216 & 3.8453 \\
    4  & \textbf{0.0111} & \textbf{0.3449} & 0.0113 & 0.4316 & 0.0321 & 2.3263 & 0.0113 & 0.7295 & 0.0115 & 1.7164 & 0.0118 & 147.5734 & 0.0132 & 0.8486          & 0.0145          & 13.9872 & 0.0230 & 3.2335 \\
    5  & 0.0089 & 0.7682 & \textbf{0.0089} & \textbf{0.7517} & 0.0165 & 2.7514 & 0.0105 & 1.1542 & 0.0121 & 1.4956 & 0.0097 & 146.7484 & 0.0093 & 0.9220          & 0.0100          & 13.2564 & 0.0302 & 4.2646 \\
    6  & \textbf{0.0064} & \textbf{0.4580} & 0.0068 & 0.4665 & 0.0242 & 2.3640 & 0.0071 & 0.7682 & 0.0178 & 1.2518 & 0.0067 & 152.7050 & 0.0139 & 1.0683          & 0.0157          & 20.4326 & 0.0182 & 3.6185 \\
    7  & \textbf{0.0020} & 0.6738 & \textbf{0.0020} & \textbf{0.5396} & 0.0242 & 2.4556 & 0.0031 & 0.9262 & 0.0074 & 1.0400 & 0.0023 & 98.1178  & 0.0086 & 1.0572          & 0.0035          & 8.3289  & 0.0162 & 2.6615 \\
    8  & 0.0085 & 0.6416 & 0.0085 & \textbf{0.4942} & 0.0174 & 2.7398 & \textbf{0.0059} & 0.9074 & 0.0093 & 1.0095 & 0.0070 & 82.6670  & 0.0090 & 1.3706          & 0.0100          & 9.9994  & 0.0198 & 3.9372 \\
    9  & 0.0061 & \textbf{0.4423} & 0.0061 & 0.5692 & 0.0290 & 2.7755 & 0.0081 & 0.6864 & 0.0059 & 1.5081 & \textbf{0.0057} & 160.7822 & 0.0080 & 0.4597          & 0.0103          & 6.1267  & 0.0346 & 4.0015 \\
    10 & 0.0039 & 0.6513 & 0.0042 & 0.6772 & 0.0247 & 2.8280 & 0.0066 & 0.9168 & 0.0077 & 2.2245 & \textbf{0.0037} & 150.7018 & 0.0086 & 1.5702          & 0.0105          & 9.7792  & 0.0317 & 4.2307 \\
    \bottomrule
  \end{tabular}%
  }
\end{table*}

As can be seen in Tables~\ref{tab:dte_dre_comparison} and ~\ref{tab:dte_dre_kitti}, the 3xGCN+GAT model consistently achieves the lowest DTE and DRE, with GAT+2xGCN performing competitively. Since DTE and DRE measure robustly aligned translation and rotation residuals, respectively, lower values indicate reduced bias and variance in the inlier pose distribution. In regression-based VO, this directly corresponds to improved geometric consistency rather than merely fewer catastrophic failures. Mathematically, if the predicted pose can be decomposed as $\hat{\mathbf{T}} = \mathbf{T} \exp(\boldsymbol{\xi})$ with perturbation $\boldsymbol{\xi} \in SE(3)$, DTE and DRE effectively quantify the expected magnitude of translational and rotational components of $\boldsymbol{\xi}$ after robust alignment. The observed improvements therefore indicate that deeper graph aggregation reduces both systematic bias and dispersion in the learned pose increments. 

It is important to note that DRE can exceed $100^\circ$ in certain cases, since rotation error is computed as the geodesic distance on $\mathrm{SO}(3)$, it is not restricted to small perturbations and can take large values when the estimated rotation is significantly misaligned with GT. Such high errors typically arise in challenging scenarios, including wide-baseline settings or when reliable geometric correspondences are sparse, leading to incorrect estimation of the camera's orientation. On the ETH3D dataset, where vegetation and irregular depth induce noisy correspondences, the advantage of 3xGCN+GAT is more pronounced, suggesting improved robustness to geometric ambiguity. On KiTTi (consecutive frames), motion is small and well-constrained; here, improvements reflect better suppression of subtle systematic drift.

\subsection{Comparison with State-of-the-Art models}

Relative and absolute pose metrics are presented for both CNN- and graph-based models using the datasets described, first for consecutive frames in all datasets and next, for different spacing in the KiTTi sequences. For the KiTTi dataset, the evolution of camera pose in 3-D Cartesian space is also presented; the first 100 images of Sequences 01 and 09 are used. It is remarked that in contrast with graph-based models, the CNN-based models supervise Pixel Flow, Edges Flow, Temporal disparity, and are independent of camera intrinsics. Additionally, DiffPoseNet incorporates optical flow cues, to guide relative pose estimation.

\subsubsection{Consecutive frames}

For CNN-based models, input images are resized to a fixed resolution ($224\times224\times3$) and trained in a supervised manner using a regression loss defined over rotation and translation vectors, typically optimized via mean squared error (MSE). In contrast, graph-based models operate on higher-resolution inputs ($640\times480$) to preserve fine-grained correspondence information. 


\begin{table*}[htpb!]
\centering
\caption{KITTI sequences 01, 05, and 09 with consecutive frames.}
\label{tab:kitti_results_transposed1}
\resizebox{0.99\textwidth}{!}{%
\begin{tabular}{lccccc|cccc|cccc}
\toprule
\multicolumn{9}{c}{\textbf{Seq 01}} \\
\midrule
\textbf{Metric} & \textbf{RANSAC} & \textbf{MAGSAC} & \textbf{GC-RANSAC} & \textbf{MAGSAC++} & \textbf{DSAC} & \textbf{PoseNet} & \textbf{RPNet} & \textbf{RPNet+} & \textbf{DiffPoseNet (c)} & \textbf{GAT+2xGCN} & \textbf{3xGCN+GAT} & \textbf{GIN\_SumPool} & \textbf{CrossGraph} \\
\midrule
ATE (m) & 9.219 & 22.12 & 18.021 & 22.665 & 11.547 & 1.538 & 1.435 & 1.927 & 1.392 & 3.246 & \textbf{0.390} & 1.871 & 3.499 \\
APE (m) & 7.105 & 16.838 & 13.793 & 17.110 & 11.505 & 1.467 & 1.375 & 1.754 & 1.343 & 2.640 & \textbf{0.296} & 1.656 & 3.142 \\
DTE (deg) & 0.038 & 0.059 & 0.047 & 0.052 & 0.041 & 0.030 & \textbf{0.023} & 0.035 & 0.026 & 0.062 & 0.025 & \textbf{0.023} & 0.098 \\
DRE (deg) & 2.0609 & 3.779 & 3.528 & 6.097 & 3.87 & 1.708 & 122.310 & 0.894 & 9.967 & 1.263 & \textbf{0.201} & 1.652 & 126.010 \\
\midrule
\multicolumn{9}{c}{\textbf{Seq 05}} \\
\midrule
ATE (m) & 1.305 & 0.981 & 1.561 & 1.702 & 1.691 & 0.932 & 0.954 & 1.323 & 0.954 & 3.185 & \textbf{0.426} & 0.477 & 2.466 \\
APE (m) & 1.047 & 0.7536 & 1.261 & 1.336 & 1.256 & 0.908 & 0.933 & 1.261 & 1.129 & 2.563 & \textbf{0.370} & 0.454 & 2.233 \\
DTE (deg) & 0.031 & 0.032 & 0.047 & 0.051 & 0.055 & \textbf{0.016} & 0.017 & 0.020 & 0.031 & 0.071 & 0.023 & 0.021 & 0.065 \\
DRE (deg) & 2.658 & 1.626 & 1.828 & 2.178 & 2.148 & 1.263 & 137.190 & 1.047 & 9.637 & 0.223 & \textbf{0.187} & 0.582 & 124.910 \\
\midrule
\multicolumn{9}{c}{\textbf{Seq 09}} \\
\midrule
ATE (m) & 10.236 & 9.242 & 15.560 & 12.388 & 11.413 & 0.656 & 0.845 & 0.712 & 1.150 & 2.278 & \textbf{0.255} & 0.370 & 2.230 \\
APE (m) & 7.301 & 6.762 & 11.038 & 8.758 & 8.077 & 0.623 & 0.800 & 0.675 & 1.051 & 1.813 & \textbf{0.205} & 0.332 & 1.977 \\
DTE (deg) & 0.038 & 0.049 & 0.062 & 0.050 & 0.042 & 0.013 & 0.015 & 0.014 & 0.015 & 0.050 & 0.014 & \textbf{0.013} & 0.062 \\
DRE (deg) & 5.169 & 3.421 & 6.751 & 6.505 & 4.850 & 1.187 & 117.140 & 1.236 & 8.577 & 0.507 & \textbf{0.202} & 0.772 & 124.920 \\
\midrule
\bottomrule
\end{tabular}
}
\end{table*}


The results for KITTI sequences, Table~\ref{tab:kitti_results_transposed1}, indicate that graph-based models generally achieve stronger geometric consistency by explicitly modeling epipolar constraints. However, in sequences with large viewpoint changes, narrow baselines, or correspondence degradation, errors in essential matrix estimation can introduce erroneous edges during graph sparsification. GAT/GIN models may then suffer from corrupted neighborhood aggregation, leading to increased DTE/DRE and trajectory drift. Compared with sampling-based consensus (SAC) methods, including RANSAC, MAGSAC, GC-RANSAC, MAGSAC++, and DSAC, the proposed 3xGCN+GAT achieves substantially lower ATE and APE across all three sequences, although SAC methods remain competitive on individual relative translation metrics. These findings highlight the benefits of learned correspondence representations for absolute trajectory estimation while indicating sensitivity to geometric and correspondence quality. For Sequences 01 and 09, the evolution of the estimated and GT camera poses in 3-D Cartesian space are presented in Fig.~\ref{fig:consec-frame-test}. As quantitatively supported by the results in Table~\ref{tab:kitti_results_transposed1}, the 3xGCN + GAT architecture and the GIN\_SumPool module achieve the lowest ATE and APE values, producing trajectories that most closely align with the ground truth across both sequences.

\begin{table*}[htpb!]
\centering
\caption{TA dataset (tartan\_left).}
\label{tab:tartan_finetuning1}
\resizebox{0.99\textwidth}{!}{%
\begin{tabular}{lcccc|cccc}
\toprule
\textbf{Metric} & \textbf{PoseNet} & \textbf{RPNet} & \textbf{RPNet+} & \textbf{DiffPoseNet} & \textbf{GAT+2xGCN} & \textbf{3xGCN+GAT} & \textbf{GIN\_SumPool} & \textbf{CrossGraph} \\
\midrule
ATE (m) & 21.065 & 17.571 & 12.108 & 36.714 & \textbf{9.621} & 10.027 & 9.864 & 13.164 \\
APE (m) & 16.478 & 15.917 & 11.380 & 33.651 & 7.740 & \textbf{7.402} & 8.312 & 10.258 \\
DTE (deg) & 12.350 & 15.194 & 10.439 & 18.956 & 5.645 & \textbf{4.027} & 5.102 & 9.071 \\
DRE (deg) & 0.0280  & 0.0290  & 0.0250 & 0.0280  & 0.0270 & 0.0220 & 0.0190 & 0.030 \\
\bottomrule
\end{tabular}
}
\end{table*}


\begin{table*}[htpb!]
\centering
\caption{KC dataset.}
\label{tab:kingcollege-test1}
\resizebox{0.99\textwidth}{!}{%
\begin{tabular}{lcccc|cccc}
\toprule
\multicolumn{9}{c}{\textbf{Seq 02}} \\
\midrule
\textbf{Metric} & \textbf{PoseNet} & \textbf{RPNet} & \textbf{RPNet+} & \textbf{DiffPoseNet} & \textbf{GAT+2xGCN} & \textbf{3xGCN+GAT} & \textbf{GIN\_SumPool} & \textbf{CrossGraph} \\
\midrule
ATE (m) & 35.786 & 45.123 & 53.460 & 28.198 & 25.631 & 29.552 & 23.276 & \textbf{22.198} \\
APE (m) & 30.727 & 38.598 & 45.887 & 23.732 & 19.011 & 23.276 & 20.521 & \textbf{18.732} \\
DTE (deg) & 25.858 & 25.828 & 24.907 & 20.939 & 23.733 & 24.997 & 19.561 & \textbf{18.939} \\
DRE (deg) & 49.665 & 88.968 & 92.330 & 137.472 & 48.489 & 50.192 & 48.830 & \textbf{43.472} \\
\midrule
\multicolumn{9}{c}{\textbf{Seq 03}} \\
\midrule
ATE (m) & 58.577 & 77.059 & 45.059 & 140.615 & 43.922 & 45.264 & \textbf{35.219} & 37.615 \\
APE (m) & 50.571 & 66.503& 38.984 & 120.903 & 40.373 & 41.201 & \textbf{29.806} & 31.903 \\
DTE (deg) & 59.627 & 90.328 & 41.974 & 120.291 & 53.573 & 56.104 & \textbf{37.305} & 40.291 \\
DRE (deg) & 64.699 & 91.611 & 80.001 & 138.758 & 68.110 & 71.783 & \textbf{37.305} & 52.758 \\
\midrule
\multicolumn{9}{c}{\textbf{Seq 07}} \\
\midrule
ATE (m) & 28.832 & 42.538 & 34.290 & 48.092 & 22.470 & 27.686 & 19.291 & \textbf{18.092} \\
APE (m) & 24.400 & 36.273 & 29.351 & 41.321 & 20.438 & 24.165 & 17.286 & \textbf{15.321} \\
DTE (deg) & 49.597 & 57.149 & 50.057 & 52.564 & 44.729 & 47.027 & 37.983 & \textbf{35.564} \\
DRE (deg) & 24.117 & 87.530 & 28.041 & 139.854 & 23.234 & 28.441 & 22.624 & \textbf{21.854} \\
\bottomrule
\end{tabular}
}
\end{table*}


Table~\ref{tab:tartan_finetuning1} reports fine-tuning results on the TA dataset. As is evident, graph-based models, particularly GAT-2xGCN and 3xGCN\_GAT, achieve the lowest translation and rotation errors, indicating improved global pose consistency over baseline methods like PoseNet and RPNet. Table~\ref{tab:kingcollege-test1} presents fine-tuning test results on the KC dataset across sequences 02, 03, and 07. Graph-based architectures, particularly GIN\_SumPool and Cross-GAT, consistently achieve lower translation and rotation errors across sequences. This is since GNN architectures such as GCN and GIN are permutation equivariant and explicitly model relation geometry between landmarks in contrast to baseline CNN models, PoseNet and RPNet, which rely on translation-invariant spatial hierarchies. On the other hand, following the results for the ETH3D dataset listed in Table~\ref{tab:eth3d_results_models1}, certain CNN-based pose regression models outperform graph-based methods in ATE/APE on challenging indoor scenes due to their ability to implicitly learn dense pixel flow, edge flow, and temporal disparity cues, making them less sensitive to intrinsic calibration variations and sparse correspondence noise. Fewer number of epipolar matches affect the graph construction process introducing noisy poses as output.


\begin{table*}[htpb!]
\centering
\caption{ETH3D datasets.}
\label{tab:eth3d_results_models1}
\resizebox{0.99\textwidth}{!}{%
\begin{tabular}{lcccc|cccc}
\toprule
\multicolumn{9}{c}{\textbf{Botanical Garden}} \\
\midrule
\textbf{Metric} & \textbf{PoseNet} & \textbf{RPNet} & \textbf{RPNet+} & \textbf{DiffPoseNet} & \textbf{GAT+2xGCN} & \textbf{3xGCN+GAT} & \textbf{GIN\_SumPool} & \textbf{CrossGraph} \\
\midrule
ATE (m) & 2.116 & 2.012 & 1.829 & 2.560 & 1.877 & 1.820 & 1.980 & \textbf{1.818} \\
APE (m) & 2.024 & 1.996 & 1.645 & 2.299 & 1.809 & 1.619 & 1.881 & \textbf{1.606} \\
DTE (deg) & 4.093 & \textbf{2.599} & 3.171 & 2.787 & 2.911 & 3.155 & 2.863 & 3.659 \\
DRE (deg) & 70.129 & 84.264 & 88.262 & 92.868 & \textbf{52.373} & 84.983 & 83.037 & 95.402 \\
\midrule
\multicolumn{9}{c}{\textbf{Boulders}} \\
\midrule
ATE (m) & 1.842 & 1.724 & 1.460 & 2.366 & 1.359 & 1.334 & 1.390 & \textbf{1.314} \\
APE (m) & 1.627 & 1.593 & 1.309 & 2.089 & 1.210 & 1.181 & 1.150 & \textbf{1.109} \\
DTE (deg) & 2.817 & 2.251 & 2.606 & 3.107 & 2.338 & 2.660 & \textbf{1.869} & 2.091 \\
DRE (deg) & 96.246 & 61.245 & 106.690 & 63.914 & 69.266 & 59.376 & \textbf{51.344} & 101.196 \\
\midrule
\multicolumn{9}{c}{\textbf{Statue}} \\
\midrule
ATE (m) & 0.163 & 0.488 & 0.256 & 0.376 & \textbf{0.157} & 0.554 & 0.808 & 0.450 \\
APE (m) & 0.1499 & 0.272 & 0.231 & 0.304 & \textbf{0.136} & 0.492 & 0.785 & 0.399 \\
DTE (deg) & 1.836 & 1.572 & 2.033 & 2.141 & 1.869 & 2.253 & \textbf{1.553} & 2.072 \\
DRE (deg) & \textbf{36.645} & 38.088 & 70.722 & 64.040 & 55.180 & 42.976 & 92.286 & 76.703 \\
\bottomrule
\end{tabular}
}
\end{table*}


 The qualitative results further reinforce that graph-based learning architectures capture the underlying geometry of motion more effectively than CNN-based pose regression methods, particularly under varying motion dynamics. Additionally, the interpretability of the graph representations reveals that the clustered regions observed in the t-SNE embeddings correspond to selectively weighted and geometrically consistent keypoints. These clusters play a crucial role in guiding the regression process, enabling more reliable pose estimation through structured and informed feature aggregation.

\begin{figure}[htbp!]
    \centering
    \begin{tikzpicture}
        \begin{groupplot}[
            group style={
                group size=2 by 1,
                horizontal sep=0.8cm,
                vertical sep=2cm
            },
            width=4.4cm, 
            height=4.4cm, 
            grid=major,
            every axis plot/.append style={
                mark=*,
                mark size=0.8pt
            },
            legend style={font=\small, at={(0.5,-0.35)}, anchor=north, legend columns=1},
            xlabel={$x$},
            ylabel={$y$},
            zlabel={$z$},
            label style={font=\small},
            tick label style={font=\scriptsize},
            view={-10}{45} 
        ]

        \nextgroupplot[]

        \addplot3[color=black, dashed, line width=1.0pt, mark=none] coordinates {
            (0.0000,0.0000,0.0000)(0.0515,-0.0242,1.0007)(0.0497,-0.0218,0.9984)(0.0506,-0.0199,1.0001)(0.0416,-0.0199,0.9931)
            (0.0489,-0.0225,0.9972)(0.0479,-0.0248,0.9911)(0.0464,-0.0239,0.9901)(0.0524,-0.0201,0.9860)(0.0461,-0.0208,0.9863)
            (0.0443,-0.0238,0.9839)(0.0435,-0.0253,0.9807)(0.0431,-0.0265,0.9788)(0.0428,-0.0236,0.9753)(0.0488,-0.0236,0.9731)
            (0.0425,-0.0229,0.9660)(0.0503,-0.0223,0.9668)(0.0472,-0.0220,0.9638)(0.0409,-0.0221,0.9599)(0.0475,-0.0176,0.9547)
            (0.0423,-0.0193,0.9519)(0.0466,-0.0187,0.9507)(0.0494,-0.0148,0.9514)(0.0459,-0.0122,0.9515)(0.0493,-0.0162,0.9592)
            (0.0522,-0.0186,0.9620)(0.0471,-0.0140,0.9640)(0.0495,-0.0140,0.9758)(0.0477,-0.0169,0.9860)(0.0488,-0.0187,0.9918)
            (0.0441,-0.0140,0.9920)(0.0456,-0.0135,0.9997)(0.0496,-0.0150,1.0144)(0.0494,-0.0154,1.0259)(0.0467,-0.0167,1.0384)
            (0.0469,-0.0148,1.0519)(0.0523,-0.0130,1.0631)(0.0538,-0.0126,1.0765)(0.0563,-0.0125,1.0910)(0.0582,-0.0135,1.1026)
            (0.0597,-0.0151,1.1147)(0.0582,-0.0157,1.1280)(0.0570,-0.0173,1.1339)(0.0534,-0.0182,1.1413)(0.0519,-0.0183,1.1482)
            (0.0535,-0.0148,1.1603)(0.0548,-0.0110,1.1788)(0.0562,-0.0090,1.2024)(0.0539,-0.0092,1.2268)(0.0538,-0.0093,1.2508)
            (0.0552,-0.0091,1.2682)(0.0538,-0.0075,1.2938)(0.0550,-0.0089,1.3148)(0.0542,-0.0112,1.3403)(0.0554,-0.0126,1.3638)
            (0.0545,-0.0123,1.3829)(0.0564,-0.0101,1.4038)(0.0552,-0.0108,1.4234)(0.0578,-0.0095,1.4471)(0.0559,-0.0117,1.4628)
            (0.0561,-0.0122,1.4857)(0.0561,-0.0124,1.5094)(0.0548,-0.0153,1.5289)(0.0566,-0.0133,1.5507)(0.0547,-0.0113,1.5679)
            (0.0555,-0.0132,1.5971)(0.0543,-0.0139,1.6139)(0.0569,-0.0171,1.6338)(0.0577,-0.0201,1.6493)(0.0564,-0.0185,1.6586)
            (0.0589,-0.0177,1.6760)(0.0542,-0.0207,1.6894)(0.0557,-0.0214,1.7038)(0.0530,-0.0222,1.7247)(0.0569,-0.0202,1.7371)
            (0.0544,-0.0211,1.7541)(0.0554,-0.0201,1.7703)(0.0530,-0.0205,1.7805)(0.0526,-0.0221,1.7951)(0.0522,-0.0222,1.8095)
            (0.0508,-0.0229,1.8308)(0.0534,-0.0222,1.8420)(0.0513,-0.0224,1.8591)(0.0547,-0.0198,1.8705)(0.0514,-0.0182,1.8895)
            (0.0521,-0.0186,1.9025)(0.0475,-0.0245,1.9139)(0.0471,-0.0272,1.9242)(0.0479,-0.0263,1.9298)(0.0503,-0.0246,1.9439)
            (0.0502,-0.0257,1.9524)(0.0487,-0.0272,1.9667)(0.0465,-0.0282,1.9781)(0.0441,-0.0273,1.9887)(0.0457,-0.0280,2.0000)
            (0.0462,-0.0277,2.0079)(0.0484,-0.0263,2.0150)(0.0461,-0.0255,2.0250)(0.0491,-0.0283,2.0404)(0.0467,-0.0285,2.0485)
            (0.0469,-0.0266,2.0533)
        };
        \addlegendentry{GT}

        \addplot3[color=blue, mark=square*, mark size=0.5pt] coordinates {
            (0.0000,0.0000,0.0000)(-0.0002,-0.0007,0.0001)(-0.0010,-0.0010,0.0011)(-0.0015,-0.0007,0.0010)(-0.0019,-0.0010,0.0004)
            (-0.0028,-0.0005,0.0005)(-0.0034,-0.0006,0.0004)(-0.0035,-0.0003,0.0005)(-0.0034,-0.0008,0.0020)(-0.0035,-0.0009,0.0013)
            (-0.0044,-0.0005,0.0012)(-0.0047,-0.0012,0.0019)(-0.0055,-0.0013,0.0006)(-0.0054,-0.0004,0.0011)(-0.0052,0.0014,0.0008)
            (-0.0050,0.0018,0.0003)(-0.0059,0.0017,-0.0003)(-0.0058,0.0029,-0.0008)(-0.0046,0.0028,-0.0011)(-0.0045,0.0028,-0.0012)
            (-0.0046,0.0029,-0.0012)(-0.0050,0.0033,-0.0007)(-0.0056,0.0040,0.0005)(-0.0059,0.0048,0.0004)(-0.0051,0.0055,0.0007)
            (-0.0052,0.0052,0.0008)(-0.0052,0.0056,0.0006)(-0.0057,0.0058,0.0006)(-0.0057,0.0059,0.0008)(-0.0060,0.0054,0.0000)
            (-0.0060,0.0061,0.0010)(-0.0059,0.0075,0.0003)(-0.0055,0.0076,0.0003)(-0.0055,0.0068,-0.0002)(-0.0062,0.0065,-0.0007)
            (-0.0064,0.0069,-0.0012)(-0.0070,0.0069,-0.0012)(-0.0074,0.0066,-0.0014)(-0.0082,0.0062,-0.0015)(-0.0087,0.0053,-0.0013)
            (-0.0082,0.0059,-0.0010)(-0.0090,0.0055,-0.0012)(-0.0094,0.0057,-0.0013)(-0.0093,0.0062,-0.0006)(-0.0091,0.0058,-0.0010)
            (-0.0091,0.0055,-0.0010)(-0.0091,0.0049,-0.0006)(-0.0092,0.0037,-0.0005)(-0.0099,0.0042,-0.0002)(-0.0104,0.0037,0.0004)
            (-0.0110,0.0039,-0.0005)(-0.0109,0.0044,-0.0011)(-0.0108,0.0051,-0.0018)(-0.0109,0.0059,-0.0029)(-0.0114,0.0051,-0.0036)
            (-0.0110,0.0046,-0.0052)(-0.0116,0.0039,-0.0050)(-0.0126,0.0039,-0.0050)(-0.0113,0.0045,-0.0054)(-0.0110,0.0042,-0.0073)
            (-0.0107,0.0044,-0.0072)(-0.0099,0.0048,-0.0073)(-0.0097,0.0043,-0.0070)(-0.0089,0.0036,-0.0072)(-0.0097,0.0037,-0.0068)
            (-0.0101,0.0050,-0.0082)(-0.0109,0.0056,-0.0078)(-0.0111,0.0063,-0.0084)(-0.0107,0.0063,-0.0080)(-0.0102,0.0054,-0.0076)
            (-0.0103,0.0041,-0.0074)(-0.0110,0.0040,-0.0079)(-0.0101,0.0030,-0.0075)(-0.0089,0.0030,-0.0083)(-0.0090,0.0027,-0.0072)
            (-0.0099,0.0032,-0.0082)(-0.0098,0.0024,-0.0073)(-0.0095,0.0022,-0.0073)(-0.0089,0.0026,-0.0072)(-0.0075,0.0029,-0.0075)
            (-0.0069,0.0023,-0.0074)(-0.0068,0.0033,-0.0068)(-0.0083,0.0038,-0.0068)(-0.0090,0.0039,-0.0073)(-0.0106,0.0041,-0.0083)
            (-0.0109,0.0048,-0.0081)(-0.0115,0.0047,-0.0070)(-0.0111,0.0063,-0.0062)(-0.0119,0.0072,-0.0061)(-0.0115,0.0071,-0.0054)
            (-0.0108,0.0077,-0.0053)(-0.0104,0.0060,-0.0050)(-0.0104,0.0063,-0.0057)(-0.0116,0.0057,-0.0049)(-0.0115,0.0069,-0.0059)
            (-0.0114,0.0064,-0.0071)(-0.0113,0.0057,-0.0054)(-0.0113,0.0053,-0.0067)(-0.0121,0.0046,-0.0068)(-0.0112,0.0044,-0.0073)
            (-0.0117,0.0044,-0.0063)
        };
        \addlegendentry{RPNet}

        \addplot3[color=teal, mark=*, mark size=0.5pt] coordinates {
            (0.0000,0.0000,0.0000)(-0.0001,-0.0005,0.0252)(-0.0002,-0.0011,0.0505)(-0.0003,-0.0016,0.0758)(-0.0003,-0.0022,0.1011)
            (-0.0003,-0.0028,0.1262)(-0.0003,-0.0034,0.1513)(-0.0003,-0.0040,0.1765)(-0.0002,-0.0046,0.2015)(-0.0000,-0.0052,0.2265)
            (0.0001,-0.0058,0.2515)(0.0003,-0.0065,0.2764)(0.0006,-0.0071,0.3012)(0.0008,-0.0077,0.3261)(0.0011,-0.0083,0.3511)
            (0.0015,-0.0090,0.3760)(0.0018,-0.0096,0.4009)(0.0022,-0.0102,0.4257)(0.0027,-0.0108,0.4505)(0.0031,-0.0114,0.4753)
            (0.0036,-0.0120,0.5000)(0.0042,-0.0126,0.5249)(0.0047,-0.0132,0.5498)(0.0053,-0.0138,0.5747)(0.0059,-0.0144,0.5995)
            (0.0066,-0.0150,0.6244)(0.0072,-0.0156,0.6493)(0.0079,-0.0162,0.6743)(0.0087,-0.0168,0.6993)(0.0094,-0.0174,0.7245)
            (0.0102,-0.0180,0.7497)(0.0110,-0.0187,0.7749)(0.0119,-0.0193,0.8001)(0.0128,-0.0200,0.8253)(0.0137,-0.0207,0.8505)
            (0.0147,-0.0214,0.8757)(0.0157,-0.0221,0.9009)(0.0168,-0.0228,0.9262)(0.0178,-0.0236,0.9515)(0.0190,-0.0243,0.9768)
            (0.0201,-0.0251,1.0022)(0.0213,-0.0260,1.0276)(0.0225,-0.0268,1.0531)(0.0238,-0.0277,1.0785)(0.0251,-0.0286,1.1041)
            (0.0264,-0.0296,1.1297)(0.0278,-0.0306,1.1554)(0.0292,-0.0316,1.1812)(0.0307,-0.0327,1.2068)(0.0322,-0.0338,1.2325)
            (0.0337,-0.0350,1.2583)(0.0353,-0.0362,1.2840)(0.0369,-0.0375,1.3098)(0.0385,-0.0388,1.3355)(0.0402,-0.0402,1.3613)
            (0.0419,-0.0415,1.3870)(0.0436,-0.0430,1.4127)(0.0454,-0.0445,1.4385)(0.0472,-0.0460,1.4643)(0.0490,-0.0476,1.4901)
            (0.0509,-0.0492,1.5158)(0.0528,-0.0509,1.5415)(0.0547,-0.0526,1.5673)(0.0567,-0.0544,1.5930)(0.0587,-0.0562,1.6188)
            (0.0607,-0.0581,1.6445)(0.0628,-0.0600,1.6703)(0.0649,-0.0619,1.6960)(0.0670,-0.0639,1.7216)(0.0692,-0.0659,1.7472)
            (0.0714,-0.0680,1.7727)(0.0736,-0.0701,1.7981)(0.0759,-0.0722,1.8237)(0.0781,-0.0744,1.8492)(0.0805,-0.0766,1.8747)
            (0.0828,-0.0788,1.9002)(0.0852,-0.0811,1.9257)(0.0877,-0.0834,1.9511)(0.0901,-0.0857,1.9765)(0.0926,-0.0881,2.0019)
            (0.0952,-0.0905,2.0273)(0.0977,-0.0930,2.0527)(0.1003,-0.0955,2.0780)(0.1030,-0.0980,2.1033)(0.1056,-0.1005,2.1285)
            (0.1083,-0.1030,2.1537)(0.1111,-0.1056,2.1789)(0.1138,-0.1082,2.2040)(0.1166,-0.1108,2.2291)(0.1195,-0.1134,2.2542)
            (0.1223,-0.1161,2.2791)(0.1252,-0.1187,2.3043)(0.1282,-0.1215,2.3294)(0.1312,-0.1242,2.3545)(0.1342,-0.1269,2.3796)
            (0.1372,-0.1297,2.4048)(0.1403,-0.1325,2.4300)(0.1434,-0.1354,2.4551)(0.1466,-0.1382,2.4803)(0.1497,-0.1411,2.5052)
            (0.1529,-0.1440,2.5301)
        };
        \addlegendentry{3xGCN+GAT}

        \nextgroupplot[]

        \addplot3[color=black, dashed, line width=1.0pt, mark=none] coordinates {
            (0.0000,0.0000,0.0000)(0.0214,-0.0085,0.2881)(0.0154,-0.0068,0.2928)(0.0127,-0.0047,0.2987)(0.0086,-0.0039,0.3061)
            (0.0048,-0.0029,0.3156)(0.0023,-0.0037,0.3233)(-0.0039,-0.0069,0.3375)(-0.0084,-0.0086,0.3549)(-0.0100,-0.0065,0.3594)
            (-0.0051,-0.0051,0.3729)(-0.0027,-0.0040,0.3855)(-0.0058,-0.0054,0.3942)(-0.0162,-0.0074,0.4016)(-0.0165,-0.0077,0.4163)
            (-0.0151,-0.0074,0.4278)(-0.0090,-0.0070,0.4421)(-0.0040,-0.0052,0.4527)(-0.0058,-0.0039,0.4684)(-0.0099,-0.0044,0.4816)
            (-0.0050,-0.0042,0.5022)(0.0000,-0.0051,0.5045)(0.0046,-0.0055,0.5251)(-0.0030,-0.0052,0.5356)(-0.0102,-0.0055,0.5494)
            (-0.0066,-0.0096,0.5600)(0.0044,-0.0149,0.5684)(0.0067,-0.0111,0.5805)(0.0017,-0.0076,0.5893)(-0.0026,-0.0076,0.6065)
            (0.0010,-0.0051,0.6157)(-0.0009,0.0009,0.6300)(-0.0009,0.0081,0.6425)(-0.0030,0.0053,0.6562)(-0.0034,0.0071,0.6689)
            (-0.0036,0.0025,0.6831)(-0.0134,-0.0006,0.7122)(-0.0149,-0.0036,0.6897)(-0.0176,-0.0037,0.7195)(-0.0163,-0.0030,0.7363)
            (-0.0141,-0.0047,0.7473)(-0.0080,-0.0085,0.7618)(-0.0079,-0.0087,0.7706)(-0.0080,-0.0086,0.7878)(-0.0066,-0.0114,0.7940)
            (-0.0210,-0.0217,0.8213)(-0.0180,-0.0203,0.8074)(-0.0226,-0.0236,0.8276)(-0.0155,-0.0175,0.8316)(-0.0155,-0.0156,0.8429)
            (-0.0127,-0.0130,0.8505)(-0.0160,-0.0156,0.8637)(-0.0099,-0.0147,0.8710)(-0.0104,-0.0154,0.8879)(-0.0118,-0.0152,0.8985)
            (-0.0095,-0.0183,0.9081)(-0.0084,-0.0161,0.9181)(-0.0088,-0.0117,0.9284)(-0.0110,-0.0121,0.9417)(-0.0077,-0.0133,0.9546)
            (-0.0081,-0.0139,0.9690)(-0.0084,-0.0117,0.9743)(-0.0021,-0.0134,1.0071)(0.0021,-0.0152,1.0243)(0.0051,-0.0184,1.0301)
            (0.0028,-0.0170,1.0430)(-0.0048,-0.0150,1.0544)(-0.0067,-0.0147,1.0638)(-0.0071,-0.0163,1.0694)(-0.0078,-0.0204,1.0712)
            (-0.0052,-0.0231,1.0687)(-0.0066,-0.0246,1.0674)(-0.0057,-0.0206,1.0647)(-0.0083,-0.0170,1.0760)(-0.0201,-0.0125,1.0757)
            (-0.0277,-0.0139,1.0785)(-0.0210,-0.0211,1.0743)(-0.0336,-0.0174,1.0651)(-0.0238,-0.0175,1.0611)(-0.0271,-0.0129,1.0582)
            (-0.0186,-0.0133,1.0671)(-0.0199,-0.0193,1.0706)(-0.0179,-0.0216,1.0761)(-0.0167,-0.0172,1.0754)(-0.0169,-0.0093,1.0754)
            (-0.0139,-0.0117,1.0785)(-0.0196,-0.0070,1.0731)(-0.0193,-0.0095,1.0849)(-0.0253,-0.0155,1.0728)(-0.0204,-0.0217,1.0816)
            (-0.0232,-0.0202,1.0842)(-0.0188,-0.0179,1.0903)(-0.0197,-0.0178,1.0920)(-0.0178,-0.0184,1.0949)(-0.0175,-0.0172,1.1120)
            (-0.0160,-0.0171,1.1084)(-0.0152,-0.0197,1.0953)(-0.0290,-0.0259,1.1129)(-0.0268,-0.0245,1.1154)(-0.0338,-0.0293,1.1249)
            (-0.0284,-0.0247,1.1292)
        };
        \addlegendentry{GT}

        \addplot3[color=red, mark=*, mark size=0.5pt] coordinates {
            (0.0000,0.0000,0.0000)(0.0017,0.0006,-0.0095)(-0.0001,0.0028,-0.0143)(-0.0009,0.0039,-0.0259)(-0.0008,0.0099,-0.0398)
            (0.0045,0.0094,-0.0465)(0.0079,0.0128,-0.0696)(0.0084,0.0147,-0.0779)(0.0085,0.0228,-0.0857)(0.0050,0.0269,-0.0887)
            (-0.0015,0.0289,-0.0943)(-0.0055,0.0337,-0.0877)(-0.0107,0.0371,-0.0886)(-0.0115,0.0417,-0.0897)(-0.0135,0.0490,-0.0836)
            (-0.0135,0.0504,-0.0763)(-0.0177,0.0518,-0.0672)(-0.0231,0.0511,-0.0589)(-0.0329,0.0520,-0.0551)(-0.0425,0.0507,-0.0415)
            (-0.0478,0.0535,-0.0352)(-0.0504,0.0547,-0.0232)(-0.0520,0.0494,-0.0176)(-0.0572,0.0486,-0.0150)(-0.0688,0.0473,-0.0142)
            (-0.0788,0.0474,-0.0110)(-0.0818,0.0471,-0.0112)(-0.0907,0.0477,-0.0062)(-0.0963,0.0488,-0.0044)(-0.1005,0.0502,-0.0028)
            (-0.0966,0.0506,-0.0059)(-0.0959,0.0519,-0.0105)(-0.0915,0.0553,-0.0124)(-0.0900,0.0563,-0.0163)(-0.0806,0.0646,-0.0202)
            (-0.0714,0.0746,-0.0247)(-0.0599,0.0787,-0.0309)(-0.0529,0.0859,-0.0298)(-0.0422,0.0920,-0.0321)(-0.0302,0.0990,-0.0374)
            (-0.0209,0.1057,-0.0415)(-0.0115,0.1096,-0.0415)(-0.0008,0.1164,-0.0468)(0.0085,0.1253,-0.0500)(0.0181,0.1304,-0.0562)
            (0.0324,0.1383,-0.0568)(0.0433,0.1461,-0.0576)(0.0530,0.1535,-0.0644)(0.0702,0.1563,-0.0684)(0.0818,0.1640,-0.0693)
            (0.0933,0.1702,-0.0733)(0.1057,0.1765,-0.0767)(0.1177,0.1830,-0.0783)(0.1338,0.1919,-0.0788)(0.1495,0.1968,-0.0857)
            (0.1572,0.2045,-0.0849)(0.1708,0.2055,-0.0897)(0.1858,0.2122,-0.0922)(0.1990,0.2209,-0.0906)(0.2130,0.2223,-0.0937)
            (0.2251,0.2260,-0.0912)(0.2376,0.2295,-0.0947)(0.2504,0.2355,-0.0918)(0.2635,0.2369,-0.0972)(0.2798,0.2428,-0.0978)
            (0.2960,0.2426,-0.0985)(0.3084,0.2450,-0.0931)(0.3253,0.2453,-0.0956)(0.3371,0.2514,-0.0957)(0.3510,0.2531,-0.0966)
            (0.3571,0.2586,-0.0981)(0.3625,0.2599,-0.0981)(0.3734,0.2631,-0.1010)(0.3754,0.2656,-0.0994)(0.3823,0.2702,-0.1022)
            (0.3904,0.2734,-0.1083)(0.3958,0.2753,-0.1125)(0.4011,0.2762,-0.1096)(0.4100,0.2796,-0.1150)(0.4162,0.2830,-0.1115)
            (0.4197,0.2822,-0.1116)(0.4299,0.2822,-0.1117)(0.4372,0.2845,-0.1087)(0.4479,0.2862,-0.1069)(0.4537,0.2913,-0.1023)
            (0.4589,0.2956,-0.1039)(0.4634,0.3002,-0.0977)(0.4647,0.3018,-0.0999)(0.4692,0.3112,-0.0994)(0.4654,0.3113,-0.0967)
            (0.4739,0.3104,-0.0999)(0.4794,0.3120,-0.1006)(0.4831,0.3120,-0.1001)(0.4877,0.3212,-0.0920)(0.4876,0.3238,-0.0882)
            (0.4851,0.3306,-0.0800)(0.4815,0.3356,-0.0779)(0.4806,0.3403,-0.0797)(0.4834,0.3469,-0.0778)(0.4791,0.3532,-0.0816)
            (0.4784,0.3512,-0.0778)
        };
        \addlegendentry{PoseNet}

        \addplot3[color=teal, mark=*, mark size=0.5pt] coordinates {
            (0.0000,0.0000,0.0000)(-0.0001,-0.0003,0.0169)(-0.0002,-0.0007,0.0339)(-0.0002,-0.0011,0.0510)(-0.0002,-0.0015,0.0681)
            (-0.0002,-0.0020,0.0853)(-0.0002,-0.0025,0.1025)(-0.0002,-0.0030,0.1198)(-0.0002,-0.0036,0.1372)(-0.0001,-0.0042,0.1545)
            (0.0000,-0.0049,0.1719)(0.0001,-0.0056,0.1892)(0.0003,-0.0063,0.2066)(0.0004,-0.0071,0.2239)(0.0006,-0.0079,0.2412)
            (0.0008,-0.0088,0.2584)(0.0011,-0.0097,0.2757)(0.0013,-0.0106,0.2929)(0.0016,-0.0115,0.3101)(0.0019,-0.0125,0.3273)
            (0.0022,-0.0136,0.3444)(0.0026,-0.0146,0.3616)(0.0030,-0.0157,0.3787)(0.0034,-0.0168,0.3959)(0.0038,-0.0180,0.4130)
            (0.0042,-0.0192,0.4301)(0.0046,-0.0204,0.4473)(0.0051,-0.0217,0.4644)(0.0056,-0.0230,0.4816)(0.0061,-0.0243,0.4987)
            (0.0066,-0.0256,0.5158)(0.0071,-0.0270,0.5329)(0.0077,-0.0284,0.5499)(0.0083,-0.0298,0.5670)(0.0089,-0.0313,0.5840)
            (0.0095,-0.0327,0.6010)(0.0101,-0.0343,0.6181)(0.0108,-0.0358,0.6351)(0.0115,-0.0374,0.6522)(0.0121,-0.0390,0.6692)
            (0.0128,-0.0406,0.6863)(0.0136,-0.0422,0.7033)(0.0143,-0.0439,0.7203)(0.0151,-0.0456,0.7373)(0.0158,-0.0473,0.7543)
            (0.0166,-0.0491,0.7713)(0.0174,-0.0509,0.7883)(0.0183,-0.0527,0.8053)(0.0191,-0.0545,0.8223)(0.0200,-0.0564,0.8393)
            (0.0208,-0.0583,0.8562)(0.0217,-0.0602,0.8732)(0.0226,-0.0621,0.8902)(0.0236,-0.0641,0.9071)(0.0245,-0.0661,0.9240)
            (0.0255,-0.0681,0.9410)(0.0265,-0.0701,0.9579)(0.0275,-0.0722,0.9748)(0.0285,-0.0743,0.9917)(0.0296,-0.0764,1.0086)
            (0.0306,-0.0785,1.0255)(0.0317,-0.0807,1.0424)(0.0328,-0.0829,1.0592)(0.0339,-0.0851,1.0760)(0.0350,-0.0873,1.0928)
            (0.0361,-0.0895,1.1096)(0.0373,-0.0918,1.1264)(0.0385,-0.0941,1.1432)(0.0396,-0.0964,1.1601)(0.0408,-0.0987,1.1769)
            (0.0421,-0.1010,1.1936)(0.0433,-0.1034,1.2105)(0.0445,-0.1058,1.2274)(0.0458,-0.1082,1.2444)(0.0471,-0.1107,1.2612)
            (0.0484,-0.1132,1.2781)(0.0497,-0.1157,1.2949)(0.0511,-0.1182,1.3117)(0.0524,-0.1207,1.3286)(0.0538,-0.1233,1.3454)
            (0.0552,-0.1259,1.3621)(0.0566,-0.1285,1.3789)(0.0580,-0.1311,1.3956)(0.0595,-0.1338,1.4123)(0.0610,-0.1365,1.4291)
            (0.0624,-0.1391,1.4458)(0.0639,-0.1419,1.4625)(0.0655,-0.1446,1.4792)(0.0670,-0.1473,1.4959)(0.0686,-0.1501,1.5126)
            (0.0702,-0.1529,1.5292)(0.0718,-0.1557,1.5460)(0.0734,-0.1585,1.5627)(0.0750,-0.1614,1.5795)(0.0767,-0.1643,1.5962)
            (0.0784,-0.1672,1.6129)(0.0801,-0.1702,1.6296)(0.0818,-0.1731,1.6464)(0.0836,-0.1761,1.6631)(0.0853,-0.1791,1.6798)
            (0.0871,-0.1821,1.6965)
        };
        \addlegendentry{3xGCN+GAT}

        \end{groupplot}
    \end{tikzpicture}
    \caption{3-D estimated trajectories (KiTTi) for consecutive frames. Left: Seq 01; Right: Seq 09.}
    \label{fig:consec-frame-test}
\end{figure}


\subsubsection{Varying frame spacing}

KiTTi sequences, with a temporal spacing of $(5,10)$, are termed as wide baseline testing specifically for resource constrained systems. Such scenarios are particularly challenging, as reduced overlap between frames leads to a significant decline in reliable feature correspondences and epipolar matches. Detailed results are presented for Sequences 01, 05, and 09; results are presented in Tables~\ref{tab:kitti-0p5} and \ref{tab:kitti-1p0}, respectively, and the estimated trajectories in 3-D space are shown in Figs.~\ref{fig:cart3d-wide-0p5} and \ref{fig:cart3d-wide-1p0}, respectively.

At a gap of 5 frames,, as shown in Fig.~\ref{fig:cart3d-wide-0p5}, both CNN and GNN models exhibit trajectory drift; however, clear differences emerge in their geometric consistency. The CNN-based models show larger deviations in heading direction and accumulate drift more rapidly, indicating sensitivity to appearance changes and reduced correspondence quality. In contrast, the GNN-based models maintain a trajectory that is more closely aligned with the ground truth, with comparatively lower angular deviation. This suggests that the GNN effectively leverages the remaining correspondences by enforcing relational constraints, thereby preserving epipolar consistency even under reduced overlap. The improved stability indicates that the model has learned to encode geometric relationships rather than relying solely on local appearance cues.

At the larger baseline condition, 10 frames' gap, as shown in Fig.~\ref{fig:cart3d-wide-1p0}, the separation between frames increases and overlap between frames drops significantly, making pose estimation inherently more ambiguous for both architectures. As a result, both CNN and GNN models exhibit increased drift over time. However, an important distinction remains: while CNN-based trajectories diverge unpredictably due to the lack of geometric grounding, GNN-based models tend to preserve a more consistent motion direction, even as drift accumulates. This behavior indicates that, although sparse correspondences limit the available geometric signal, the GNN continues to exploit relational structure among the remaining keypoints. The degradation in performance can therefore be attributed not to the failure of the graph formulation, but to the inherent scarcity of epipolar constraints at large baselines, which restricts the effectiveness of message passing.


\begin{table*}[htpb!]
\centering
\caption{KiTTi sequences 01, 05, and 09 with 5 frames difference}
\label{tab:kitti-0p5}
\resizebox{0.99\textwidth}{!}{%
\begin{tabular}{lcccc|cccc}
\toprule
\multicolumn{9}{c}{\textbf{Seq 01}} \\
\midrule
\textbf{Metric} & \textbf{PoseNet} & \textbf{RPNet} & \textbf{RPNet+} & \textbf{DiffPoseNet (c)} & \textbf{GAT+2xGCN} & \textbf{3xGCN+GAT} & \textbf{GIN\_SumPool} & \textbf{CrossGraph} \\
\midrule
ATE (m) & 6.217 & 4.874 & 5.333 & 4.481 & 2.281 & 4.608 & \textbf{1.384} & 25.058 \\
APE (m) & 5.301 & 4.174 & 4.598 & 3.870 & 1.815 & 3.218 & \textbf{0.915} & 21.401 \\
DTE (deg) & 2.986 & 2.338 & 2.439 & 2.201 & 2.617 & 3.589 & \textbf{1.886} & 21.470 \\
DRE (deg) & 7.703 & 127.130 & 8.229 & 16.083 & \textbf{7.593} & 7.730 & 8.893 & 99.111 \\
\midrule
\multicolumn{9}{c}{\textbf{Seq 05}} \\
\midrule
ATE (m) & 2.971 & 3.862 & 3.739 & 4.255 & 3.668 & 6.054 & \textbf{0.667} & 19.070 \\
APE (m) & 2.506 & 3.243 & 3.127 & 3.547 & 2.578 & 4.545 & \textbf{0.495} & 16.282 \\
DTE (deg) & 1.530 & 1.808 & 1.874 & 2.100 & 2.192 & 3.500 & \textbf{0.692} & 16.405 \\
DRE (deg) & 1.433 & 137.740 & 1.837 & 8.013 & 0.713 & \textbf{0.655} & 1.013 & 98.665 \\
\midrule
\multicolumn{9}{c}{\textbf{Seq 09}} \\
\midrule
ATE (m) & 2.044 & 2.757 & 2.223 & 1.122 & 1.714 & 3.882 & \textbf{0.188} & 15.820 \\
APE (m) & 1.667 & 2.222 & 1.847 & 0.945 & 1.273 & 3.053 & \textbf{0.154} & 13.493 \\
DTE (deg) & 1.084 & 1.512 & 1.267 & 0.663 & 0.998 & 2.136 & \textbf{0.108} & 13.567 \\
DRE (deg) & 2.281 & 155.780 & 2.367 & 7.788 & 1.606 & \textbf{1.530} & 1.672 & 98.350 \\
\midrule
\bottomrule
\end{tabular}
}
\end{table*}

\begin{table*}[htpb!]
\centering
\caption{KiTTi sequences 01, 05, and 09 with 10 frames difference}
\label{tab:kitti-1p0}
\resizebox{0.99\textwidth}{!}{%
\begin{tabular}{lcccc|cccc}
\toprule
\multicolumn{9}{c}{\textbf{Seq 01}} \\
\midrule
\textbf{Metric} & \textbf{PoseNet} & \textbf{RPNet} & \textbf{RPNet+} & \textbf{DiffPoseNet (c)} & \textbf{GAT+2xGCN} & \textbf{3xGCN+GAT} & \textbf{GIN\_SumPool} & \textbf{CrossGraph} \\
\midrule
ATE (m) & 12.064 & 7.645 & 10.124 & 5.635 & 5.822 & \textbf{4.687} & 6.355 & 18.601 \\
APE (m) & 8.531 & 5.406 & 7.159 & 3.985 & 4.117 & \textbf{3.314} & 4.493 & 13.153 \\
DTE (deg) & 17.061 & 10.811 & 14.318 & 7.969 & 8.234 & \textbf{6.628} & 8.987 & 26.306 \\
DRE (deg) & \textbf{22.405} & 92.577 & 23.056 & 38.062 & 28.187 & 23.115 & 30.388 & 169.775 \\
\midrule
\multicolumn{9}{c}{\textbf{Seq 05}} \\
\midrule
ATE (m) & 1.592 & 2.545 & 2.526 & 3.160 & \textbf{0.112} & 0.405 & 0.130 & 5.513 \\
APE (m) & 1.126 & 1.800 & 1.787 & 2.234 & \textbf{0.079} & 0.286 & 0.092 & 3.898 \\
DTE (deg) & 2.251 & 3.600 & 3.572 & 4.468 & \textbf{0.160} & 0.573 & 0.184 & 7.797 \\
DRE (deg) & 1.286 & 176.850 & 1.830 & 10.958 & \textbf{1.275} & 1.412 & 2.139 & 149.001 \\
\midrule
\multicolumn{9}{c}{\textbf{Seq 09}} \\
\midrule
ATE (m) & 5.503 & 5.747 & 4.313 & 2.814 & 2.751 & \textbf{1.124} & 1.305 & 12.008 \\
APE (m) & 3.891 & 4.064 & 3.050 & 1.990 & 1.945 & \textbf{0.795} & 0.923 & 8.491 \\
DTE (deg) & 7.783 & 8.128 & 6.100 & 3.980 & 3.890 & \textbf{1.590} & 1.845 & 16.982 \\
DRE (deg) & 0.998 & 162.580 & \textbf{0.411} & 8.180 & 0.681 & 1.119 & 1.825 & 149.090 \\
\bottomrule
\end{tabular}
}
\end{table*}

\begin{figure}[htbp!]
    \centering
    \begin{tikzpicture}
        \begin{groupplot}[
            group style={
                group size=2 by 1,
                horizontal sep=0.9cm,
                vertical sep=2cm
            },
            width=4.4cm, 
            height=4.4cm,
            grid=major,
            every axis plot/.append style={
                mark=*,
                mark size=0.8pt
            },
            legend style={
                font=\small, 
                at={(0.5,-0.35)}, 
                anchor=north, 
                legend columns=1
            },
            xlabel={$x$},
            ylabel={$y$},
            zlabel={$z$},
            label style={font=\small},
            tick label style={font=\scriptsize},
            view={-10}{45} 
        ]

        \nextgroupplot[]

        \addplot3[color=black, dashed, line width=1.0pt, mark=none] coordinates {
            (0.0000,0.0000,0.0000)(0.6972,-0.1439,4.9346)(0.6820,-0.1594,4.8837)(0.6400,-0.1677,4.8296)
            (0.6257,-0.1413,4.7553)(0.6689,-0.1235,4.7161)(0.7782,-0.1274,4.8241)(0.8634,-0.1342,5.0297)
            (1.0596,-0.1201,5.3142)(1.1711,-0.1275,5.5659)(1.2715,-0.0873,5.9741)(1.3378,-0.1291,6.5396)
            (1.3892,-0.1049,7.0664)(1.4024,-0.1237,7.5975)(1.4020,-0.1490,8.0749)(1.3791,-0.1771,8.4523)
            (1.3485,-0.1708,8.8296)(1.3199,-0.1607,9.2074)(1.2664,-0.1740,9.5090)(1.2465,-0.2168,9.7865)
        };
        \addlegendentry{GT}

        \addplot3[color=green!60!black, mark=triangle*, mark size=0.5pt] coordinates {
            (0.0000,0.0000,0.0000)(0.0045,-0.0589,-0.1369)(0.0949,-0.1586,-0.1842)(0.1238,-0.2115,-0.3988)
            (0.2008,-0.2866,-0.4615)(0.1865,-0.3654,-0.4850)(0.1689,-0.4591,-0.3041)(0.2878,-0.5682,-0.3754)
            (0.5023,-0.6511,-0.5703)(0.4750,-0.6917,-0.7172)(0.5924,-0.7409,-0.9174)(0.5199,-0.7768,-0.7886)
            (0.5648,-0.8537,-0.7696)(0.5306,-0.9618,-0.4078)(0.7194,-1.1215,-0.1808)(0.7009,-1.1878,-0.1278)
            (0.9591,-1.3626,0.0450)(0.8151,-1.3687,0.0444)(0.7125,-1.4488,0.2543)(0.5559,-1.5353,0.4687)
        };
        \addlegendentry{RPNet+}

        \addplot3[color=magenta, mark=triangle*, mark size=0.5pt] coordinates {
            (0.0000,0.0000,0.0000)(0.1099,-0.0001,0.5419)(0.2888,0.0171,1.0649)(0.5277,0.0486,1.5626)
            (0.8221,0.0955,2.0283)(1.1655,0.1549,2.4576)(1.5528,0.2225,2.8465)(1.9797,0.3012,3.1890)
            (2.4201,0.3834,3.5132)(2.8603,0.4548,3.8401)(3.3059,0.5194,4.1612)(3.7604,0.5799,4.4703)
            (4.2178,0.6383,4.7754)(4.6765,0.6924,5.0794)(5.1382,0.7437,5.3795)(5.6032,0.7891,5.6752)
            (6.0725,0.8302,5.9648)(6.5477,0.8637,6.2456)(7.0354,0.8896,6.5049)(7.5365,0.9073,6.7381)
        };
        \addlegendentry{GIN\_SumPool}

        \nextgroupplot[]

        \addplot3[color=black, dashed, line width=1.0pt, mark=none] coordinates {
            (0.0000,0.0000,0.0000)(0.0405,0.0148,-0.2316)(-0.0019,0.0687,-0.3495)(-0.0214,0.0940,-0.6301)
            (-0.0195,0.2417,-0.9700)(0.1108,0.2297,-1.1322)(0.1928,0.3118,-1.6959)(0.2054,0.3591,-1.8986)
            (0.2062,0.5565,-2.0872)(0.1225,0.6563,-2.1606)(-0.0368,0.7038,-2.2984)(-0.1338,0.8220,-2.1358)
            (-0.2595,0.9033,-2.1578)(-0.2807,1.0165,-2.1851)(-0.3288,1.1933,-2.0368)(-0.3279,1.2292,-1.8585)
            (-0.4312,1.2611,-1.6365)(-0.5636,1.2443,-1.4359)(-0.8027,1.2673,-1.3419)(-1.0351,1.2347,-1.0116)
        };
        \addlegendentry{GT}

        \addplot3[color=green!60!black, mark=triangle*, mark size=0.5pt] coordinates {
            (0.0000,0.0000,0.0000)(0.0045,-0.0589,-0.1369)(0.0949,-0.1586,-0.1842)(0.1238,-0.2115,-0.3988)
            (0.2008,-0.2866,-0.4615)(0.1865,-0.3654,-0.4850)(0.1689,-0.4591,-0.3041)(0.2878,-0.5682,-0.3754)
            (0.5023,-0.6511,-0.5703)(0.4750,-0.6917,-0.7172)(0.5924,-0.7409,-0.9174)(0.5199,-0.7768,-0.7886)
            (0.5648,-0.8537,-0.7696)(0.5306,-0.9618,-0.4078)(0.7194,-1.1215,-0.1808)(0.7009,-1.1878,-0.1278)
            (0.9591,-1.3626,0.0450)(0.8151,-1.3687,0.0444)(0.7125,-1.4488,0.2543)(0.5559,-1.5353,0.4687)
        };
        \addlegendentry{RPNet+}

        \addplot3[color=magenta, mark=triangle*, mark size=0.5pt] coordinates {
            (0.0000,0.0000,0.0000)(-0.0061,-0.0099,0.3037)(-0.0750,0.0065,0.5993)(-0.1591,0.0178,0.8912)
            (-0.2334,0.0093,1.1858)(-0.3080,-0.0060,1.4800)(-0.3915,-0.0260,1.7716)(-0.4707,-0.0515,2.0639)
            (-0.5552,-0.0794,2.3546)(-0.6488,-0.1102,2.6421)(-0.7578,-0.1427,2.9240)(-0.8711,-0.1800,3.2035)
            (-0.9834,-0.2204,3.4831)(-1.0873,-0.2647,3.7652)(-1.1943,-0.3115,4.0459)(-1.3226,-0.3588,4.3173)
            (-1.4528,-0.4105,4.5871)(-1.5836,-0.4659,4.8558)(-1.7225,-0.5236,5.1199)(-1.8687,-0.5843,5.3794)
        };
        \addlegendentry{GIN\_SumPool}

        \end{groupplot}
    \end{tikzpicture}
    \caption{3-D estimated trajectories (KiTTi) with 5 frame spacing. Left: Seq 01; Right: Seq 09.}
    \label{fig:cart3d-wide-0p5}
\end{figure}

\begin{figure}[htbp]
    \centering
    \begin{tikzpicture}
        \begin{groupplot}[
            group style={
                group size=2 by 1,
                horizontal sep=0.9cm,
                vertical sep=2cm
            },
            width=4.4cm, 
            height=4.4cm,
            grid=major,
            every axis plot/.append style={
                mark=*,
                mark size=0.8pt
            },
            legend style={
                font=\small, 
                at={(0.5,-0.35)}, 
                anchor=north, 
                legend columns=1
            },
            xlabel={$x$},
            ylabel={$y$},
            zlabel={$z$},
            label style={font=\small},
            tick label style={font=\scriptsize},
            view={-10}{45} 
        ]

        \nextgroupplot[]

        \addplot3[color=black, dashed, line width=1.0pt, mark=none] coordinates {
            (0.0000,0.0000,0.0000)(0.1223,0.0968,-0.2399)(-0.0599,0.1158,-0.2331)(-1.0078,0.2156,0.4727)
            (-2.4623,0.2422,1.3282)(-3.9552,0.3536,2.9463)(-5.1514,0.3728,4.6638)(-5.8008,0.3408,6.1394)
            (-6.2197,0.3128,7.5276)(-6.4425,0.2877,8.6319)
        };
        \addlegendentry{GT}

        \addplot3[color=green!60!black, mark=triangle*, mark size=0.5pt] coordinates {
            (0.0000,0.0000,0.0000)(-0.0311,-0.0877,-0.7994)(0.0806,-0.2032,-0.5692)(0.7870,-0.5245,-1.1096)
            (-0.0612,-0.4175,-2.0794)(0.1713,-0.5826,-2.7771)(0.4572,-0.7155,-3.9165)(0.2727,-0.7665,-4.5168)
            (0.7793,-0.9854,-5.4038)(0.7391,-1.0407,-6.0095)
        };
        \addlegendentry{RPNet+}

        \addplot3[color=teal, mark=*, mark size=0.5pt] coordinates {
            (0.0000,0.0000,0.0000)(0.5280,-0.0340,1.6752)(1.5525,-0.0875,2.9886)(2.9540,-0.1695,3.8704)
            (4.4965,-0.2721,4.3443)(7.7437,-0.5102,4.9792)(11.4986,-0.7917,5.5813)(14.8880,-1.0486,6.1423)
            (18.2912,-1.3058,6.7031)(21.7700,-1.5729,7.2119)
        };
        \addlegendentry{3xGCN+GAT}

        \nextgroupplot[]

        \addplot3[color=black, dashed, line width=1.0pt, mark=none] coordinates {
            (0.0000,0.0000,0.0000)(-0.3691,0.0221,1.1327)(-0.0385,-0.0630,2.3927)(0.0629,-0.1610,3.6363)
            (-0.0488,0.0788,4.8557)(0.0191,0.1061,5.8994)(0.5000,-0.0147,7.1437)(-0.0621,-0.0490,7.4466)
            (0.1375,0.0601,7.5293)(-0.0637,-0.1186,7.8346)
        };
        \addlegendentry{GT}

        \addplot3[color=green!60!black, mark=triangle*, mark size=0.5pt] coordinates {
            (0.0000,0.0000,0.0000)(-0.3185,-0.0894,0.1132)(-0.4036,-0.2334,0.6154)(-0.3574,-0.3259,0.4953)
            (-0.2424,-0.4286,0.5036)(-0.2129,-0.4896,0.7267)(-0.1439,-0.5658,0.6894)(-0.2508,-0.6909,1.1172)
            (-0.2047,-0.8153,1.2399)(-0.1831,-0.9415,1.4758)
        };
        \addlegendentry{RPNet+}

        \addplot3[color=teal, mark=*, mark size=0.5pt] coordinates {
            (0.0000,0.0000,0.0000)(-0.2343,-0.0433,2.1236)(-0.8230,-0.0896,4.3764)(-1.4903,-0.1629,6.5200)
            (-2.2976,-0.2577,8.8052)(-3.2565,-0.3491,11.0140)(-4.3336,-0.4374,13.1302)(-5.4818,-0.5296,15.2429)
            (-6.7050,-0.6150,17.1838)(-8.0061,-0.6985,19.0168)
        };
        \addlegendentry{3xGCN+GAT}

        \end{groupplot}
    \end{tikzpicture}
    \caption{3-D estimated trajectories (KiTTi) with 10 frame spacing. Left: Seq 01; Right: Seq 09.}
    \label{fig:cart3d-wide-1p0}
\end{figure}

We present additional analyses of the proposed graph architectures next. 

\section{Validation of the Proposed Graph Representation}

In this section, we present results on spectral analysis of the proposed architectures as well as analysis on the learned representations. In the former, with the use of known measures of energy and entropy, we demonstrate how the injection of anomalies or outliers changes these metrics across the graph layers. The learned representations are analysed at the node and graph levels to evaluate how correspondence-level geometric information evolves through graph message passing and subsequent global aggregation.  

\subsection{Spectral Analysis of Learned Graph Representations}

The spectral behaviour of the learned graph representations is evaluated - using low frequency energy and spectral entropy, \cite{hamilton2020graph}, as measures - across graph layers in proposed learning modules by progressively increasing outliers in correspondences. Low-frequency energy is defined by $E_{\text{low}} = \frac{\sum\limits_{i=1}^{k} \lambda_i^2}{\sum\limits_{i=1}^{n} \lambda_i^2}$, where $\lambda_i$ are the eigenvalues of the graph Laplacian and $k$ is chosen such that $k/n \approx 0.1$. Spectral entropy, $H(\mathcal{G})$, quantifies the distribution of graph frequencies and is found as $H(\mathcal{G}) = -\sum\limits_{i=1}^{n} p_i \log(p_i), \quad p_i = \frac{\lambda_i^2}{\sum\limits_{j=1}^{n} \lambda_j^2}$. Tables~\ref{tab:low_freq_energy} and~\ref{tab:spectral_entropy} report the corresponding layer-wise measurements for outlier rates ranging from $0\%$ to $60\%$.

For clean correspondences, all architectures exhibit substantially higher low-frequency energy than under outlier injected inputs, with the corresponding spectral entropy remaining low. As the outlier rate increases, low-frequency energy decreases while spectral entropy increases across all the evaluated architectures, indicating a progressive loss of smooth graph structure under correspondence corruption. The degradation is particularly pronounced for GAT\_2xGCN, whose low-frequency energy drops rapidly with increasing outlier rates. In contrast, GIN\_SumPool and CrossGraph retain higher low-frequency energy at moderate corruption levels, although their spectral entropy also increases as the noise becomes more severe.

\begin{table*}[htbp]
\centering
\caption{Layer-wise Low-Frequency Energy across synthetic outlier injection rates ($0\%$ to $60\%$).}
\label{tab:low_freq_energy}
\resizebox{\textwidth}{!}{%
\begin{tabular}{c cccc ccc cc ccc}
\toprule
\multirow{2}{*}{\textbf{Outlier \%}} & \multicolumn{4}{c}{\textbf{3xGCN\_GAT}} & \multicolumn{3}{c}{\textbf{GAT\_2xGCN}} & \multicolumn{2}{c}{\textbf{GIN\_SumPool}} & \multicolumn{3}{c}{\textbf{CrossGraph}} \\
\cmidrule(lr){2-5} \cmidrule(lr){6-8} \cmidrule(lr){9-10} \cmidrule(lr){11-13}
 & \textbf{L1} & \textbf{L2} & \textbf{L3} & \textbf{L4} & \textbf{L1} & \textbf{L2} & \textbf{L3} & \textbf{L1} & \textbf{L2} & \textbf{L1} & \textbf{L2} & \textbf{L3} \\
\midrule
\textbf{0\%}  & 0.8533 & 0.8103 & 0.7970 & 0.7956 & 0.7184 & 0.6963 & 0.6721 & 0.9691 & 0.9113 & 0.9984 & 0.9341 & 0.8526 \\
\textbf{10\%} & 0.2036 & 0.1070 & 0.0747 & 0.0705 & 0.0336 & 0.0164 & 0.0085 & 0.4867 & 0.2476 & 0.5392 & 0.3199 & 0.2421 \\
\textbf{20\%} & 0.0176 & 0.0058 & 0.0040 & 0.0038 & 0.0303 & 0.0169 & 0.0091 & 0.2076 & 0.0932 & 0.3694 & 0.1778 & 0.0855 \\
\textbf{40\%} & 0.0114 & 0.0056 & 0.0044 & 0.0038 & 0.0071 & 0.0038 & 0.0030 & 0.0501 & 0.0167 & 0.1247 & 0.0414 & 0.0081 \\
\textbf{60\%} & 0.0111 & 0.0076 & 0.0077 & 0.0063 & 0.0102 & 0.0114 & 0.0115 & 0.0348 & 0.0105 & 0.0732 & 0.0101 & 0.0052 \\
\bottomrule
\end{tabular}%
}
\end{table*}

\begin{table*}[htbp]
\centering
\caption{Layer-wise Spectral Entropy across synthetic outlier injection rates ($0\%$ to $60\%$).}
\label{tab:spectral_entropy}
\resizebox{\textwidth}{!}{%
\begin{tabular}{c cccc ccc cc ccc}
\toprule
\multirow{2}{*}{\textbf{Outlier \%}} & \multicolumn{4}{c}{\textbf{3GCN\_GAT}} & \multicolumn{3}{c}{\textbf{GAT\_2GCN}} & \multicolumn{2}{c}{\textbf{GIN\_SumPool}} & \multicolumn{3}{c}{\textbf{CrossGraph}} \\
\cmidrule(lr){2-5} \cmidrule(lr){6-8} \cmidrule(lr){9-10} \cmidrule(lr){11-13}
 & \textbf{L1} & \textbf{L2} & \textbf{L3} & \textbf{L4} & \textbf{L1} & \textbf{L2} & \textbf{L3} & \textbf{L1} & \textbf{L2} & \textbf{L1} & \textbf{L2} & \textbf{L3} \\
\midrule
\textbf{0\%}  & 1.6127 & 2.1890 & 2.3079 & 2.3276 & 2.7154 & 2.8720 & 3.0771 & 1.0215 & 1.7191 & 0.1414 & 0.1063 & 0.0811 \\
\textbf{10\%} & 5.9618 & 6.2143 & 6.0621 & 5.9755 & 7.0643 & 6.6139 & 6.1822 & 4.3416 & 5.7815 & 0.9459 & 0.5622 & 0.1004 \\
\textbf{20\%} & 7.2540 & 6.6554 & 6.2999 & 6.1740 & 7.0236 & 6.5359 & 6.2443 & 6.1482 & 6.7213 & 1.7921 & 1.2006 & 0.7938 \\
\textbf{40\%} & 7.2607 & 6.7739 & 6.5308 & 6.4224 & 7.1024 & 6.6335 & 6.4673 & 7.1516 & 7.2842 & 2.9512 & 1.8897 & 1.6312 \\
\textbf{60\%} & 7.3010 & 6.8358 & 6.6260 & 6.5183 & 7.1004 & 6.7267 & 6.6421 & 7.2311 & 7.3569 & 4.0762 & 3.4902 & 2.9117 \\
\bottomrule
\end{tabular}%
}
\end{table*}

The proposed 3xGCN\_GAT exhibits a comparatively stable layer-wise spectral evolution under increasing correspondence corruption. Its low-frequency energy decreases progressively rather than abruptly across the graph layers, while spectral entropy increases more gradually than in the other convolution-attention architectures. This indicates that the proposed architecture retains a more stable spectral representation as correspondence quality degrades, consistent with its improved ATE and APE-R performance reported in the pose estimation experiments.

\subsection{Multi-Scale Representation Analysis}

Node-level analysis uses the representations and the non-parametric Spearman's correlation measure while the graph-level analysis evaluates the intrinsic dimensionality and latent-space organization of the pooled representations. The quantitative results are reported in Tables~\ref{tab:norm_residual_corr} and \ref{tab:global_latent_analysis}.

At the node level, Table~\ref{tab:norm_residual_corr} shows that the early layers generally exhibit negative correlations, based on the Spearman coefficient, between embedding magnitude and the Sampson residual, indicating that the learned representations retain correspondence-level geometric information during the initial stages of message passing. This relationship weakens in deeper layers for the proposed 3xGCN\_GAT architecture, consistent with a transition from local correspondence information towards higher-level relational representations.These results indicate that geometric information present at the correspondence level is transformed rather than discarded during message passing.


\begin{table}[htbp]
\centering
\caption{Node Embedding Norm vs. Residual Error Spearman Correlation across layers for each model architecture.}
\label{tab:norm_residual_corr}
\small
\begin{tabular}{ll c}
\toprule
\textbf{Model Name} & \textbf{Layer} & \textbf{Spearman} \\
\midrule
\multirow{4}{*}{\textbf{3xGCN\_GAT}} 
 & Layer 1 & -0.590  \\
 & Layer 2 & -0.411  \\
 & Layer 3 & -0.100  \\
 & Layer 4 & -0.017  \\
\midrule
\multirow{3}{*}{\textbf{GAT\_2xGCN}} 
 & Layer 1 & -0.271  \\
 & Layer 2 & \phantom{-}0.015  \\
 & Layer 3 & -0.223  \\
\midrule
\multirow{2}{*}{\textbf{GIN\_SumPool}} 
 & Layer 1 & -0.222  \\
 & Layer 2 & -0.087  \\
\midrule
\multirow{3}{*}{\textbf{CrossGraph}} 
 & Layer 1 & -0.650  \\
 & Layer 2 & -0.652  \\
 & Layer 3 & \phantom{-}0.649  \\
\bottomrule
\end{tabular}
\end{table}

At the graph level, Table~\ref{tab:global_latent_analysis} shows that the proposed 3xGCN\_GAT produces a compact latent representation, with a low effective rank and only two principal components accounting for 90\% of the variance. The learned representations exhibit structured variation with respect to relative rotation and translation magnitude, while samples from different scene categories remain distributed throughout the latent space. This suggests that the pooled representation captures motion-related information without forming strong scene-specific clusters. Overall, the node- and graph-level analyses indicate that the proposed architecture transforms local correspondence geometry into a compact graph-level representation that remains structured with respect to camera motion.

\begin{table*}[htbp]
\centering
\caption{Global Latent Space Analysis: Effective Rank and PCA Components (90\% Variance) across Pre- and Post-Pooling Embeddings.}
\label{tab:global_latent_analysis}
\small
\begin{tabular}{l cc cc}
\toprule
\multirow{2}{*}{\textbf{Model Name}} & \multicolumn{2}{c}{\textbf{Pre-Pool Embedding}} & \multicolumn{2}{c}{\textbf{Post-Pool Embedding}} \\
\cmidrule(lr){2-3} \cmidrule(lr){4-5}
 & \textbf{Effective Rank} & \textbf{PCA (90\% Var)} & \textbf{Effective Rank} & \textbf{PCA (90\% Var)} \\
\midrule
\textbf{3xGCN\_GAT}   & 4.43 / 128  & 2 & 4.43 / 128  & 2 \\
\textbf{GAT\_2xGCN}   & 9.11 / 128  & 2 & 9.11 / 128  & 2 \\
\textbf{GIN\_SumPool} & 12.36 / 64  & 4 & 12.36 / 64  & 4 \\
\textbf{CrossGraph}  & 3.40 / 128  & 2 & 16.50 / 128 & 1 \\
\bottomrule
\end{tabular}
\end{table*}

\section{Conclusions and Future Directions}

This work presents an epipolar-constrained graph representation learning framework for geometric relative pose estimation from two-view image correspondences. The proposed formulation incorporates explicit geometric constraints into graph topology by refining an initial spatial $k$-NN graph using an initial estimate of Essential Matrix and Sampson distance. Unlike conventional graph formulations that primarily rely on heuristic spatial neighbourhoods, the proposed approach uses epipolar consistency to guide graph connectivity while retaining all correspondence nodes. The resulting graph representation is passed through graph convolution and attention mechanisms to obtain an local-global embedding for direct relative pose regression, providing an alternative to conventional sampling-based consensus estimators for final pose estimation.

The experimental findings demonstrate that epipolar-guided refinement reduces graph connectivity by more than 90\% by removing geometrically inconsistent edges without discarding correspondence observations. Graph structure and spectral analyses further characterize the effects of sparsification and noisy correspondences on graph representations and information propagation across graph layers. The comparative evaluation of GCN, GAT, GIN, and Graph-based Cross Attention identifies 3GCN\_GAT as a favourable architecture for balancing local geometric propagation and attention-based feature aggregation. These observations are consistent with the improvements in relative pose and trajectory estimation obtained across KITTI, ETH3D, King's College, and TartanAir under standard and wide temporal baselines. Together, these results highlight the importance of incorporating geometric consistency into graph construction rather than relying exclusively on learned feature aggregation over heuristic graph structures.

Despite the advantages, the proposed framework has certain limitations. Its graph construction depends on an initial Essential Matrix estimate, making the resulting topology sensitive to errors in geometric initialization. Furthermore, correspondence extraction and graph-based pose estimation are currently performed as separate stages, allowing matching errors to propagate into subsequent representation learning. The present formulation also focuses on pairwise relative pose estimation without explicitly modelling long-term temporal dependencies or jointly optimizing camera poses and scene structure. Nevertheless, the proposed formulation provides a useful basis for investigating geometry-guided graph representation learning in multi-view vision, particularly where relational consistency is important for geometric estimation.

Future work will investigate jointly learning correspondence confidence, graph connectivity, and pose estimation to reduce dependence on separate geometric initialization and improve adaptability under severe matching noise. Extending the framework towards multi-view geometric reasoning could enable the integration of multiple correspondence graphs, camera poses, and structural constraints for incremental mapping and loop-closure optimization. Additionally, uncertainty-aware graph learning could improve reliability in low-texture, dynamic, and wide-baseline environments. Finally, efficient graph construction and lightweight message-passing mechanisms will be explored to reduce computational overhead and facilitate deployment in real-time visual SLAM and resource-constrained robotic systems.

\section{Statements and Declarations}

\subsection{Funding}

No specific funding was acquired for this paper. This paper is the result of the work towards earning the MS (Research) degree for Mr Prateeth Rao. He was funded by IIIT-Bangalore during his tenure as a Master's student.

\subsection{Data Availability}

All data used in this paper is from publicly available sources.\footnote{
KiTTi 2012 SLAM Benchmark: \url{https://www.cvlibs.net/datasets/kitti/eval_stereo_flow.php?benchmark=stereo}; 
King's College dataset: \url{https://www.repository.cam.ac.uk/items/53788265-cb98-42ee-b85b-7a0cbc8eddb3}; 
Tartan Air dataset: \url{https://theairlab.org/tartanair-dataset/}.
}






\bibliographystyle{elsarticle-num}

\end{document}